\documentclass{ecai} 

\usepackage{latexsym}
\usepackage{amssymb}
\usepackage{amsmath}
\usepackage{amsthm}
\usepackage{booktabs}
\usepackage{enumitem}
\usepackage[pdftex]{graphicx}
\usepackage{color}

\usepackage{algorithm}
\usepackage[noend]{algpseudocode}
\usepackage{subcaption}

\newtheorem{theorem}{Theorem}
\newtheorem{lemma}[theorem]{Lemma}
\newtheorem{corollary}[theorem]{Corollary}

\newcommand{\BibTeX}{B\kern-.05em{\sc i\kern-.025em b}\kern-.08em\TeX}

\begin{document}


\begin{frontmatter}


\paperid{1247} 


\title{Structure and Reduction of MCTS for Explainable-AI}


\author[A]{\fnms{Ronit}~\snm{Bustin}\thanks{Corresponding Author. Email: ronit.bustin@gmail.com}}
\author[B]{\fnms{Claudia V.}~\snm{Goldman}\thanks{Email: claudia.goldman@mail.huji.ac.il, research was performed while this author was affiliated with General Motors}}
\address[A]{General Motors, Technical Center Israel}
\address[B]{School of Business Administration, The Hebrew University}


\begin{abstract}
Complex sequential decision-making planning problems, covering infinite states’ space have been shown to be solvable by AlphaZero type of algorithms. Such an approach that trains a neural model while simulating projection of futures with a Monte Carlo Tree Search algorithm were shown to be applicable to real life planning problems. As such, engineers and users interacting with the resulting policy of behavior might benefit from obtaining automated explanations about these planners’ decisions offline or online.
This paper focuses on the information within the Monte Carlo Tree Search data structure. Given its construction, this information contains much of the reasoning of the sequential decision-making algorithm and is essential for its explainability. We show novel methods using information theoretic tools for the simplification and reduction of the Monte Carlo Tree Search and the extraction of information. Such information can be directly used for the construction of human understandable explanations. We show that basic explainability quantities can be calculated with limited additional computational cost, as an integrated part of the Monte Carlo Tree Search construction process. We focus on the theoretical and algorithmic aspects and provide examples of how the methods presented here can be used in the construction of human understandable explanations.   
\end{abstract}

\end{frontmatter}


\section{Introduction}\label{sec:introduction}
Complex real life planning problems, such as autonomous vehicles' behavior, can be framed as sequential decision-making problems. 
Given a general route for the vehicle, the vehicle needs to determine the specific steering angle and accelerations that result in what we would consider to be good driving. Due to the complexity of the problem, in which there are many relevant entities with probabilistic behaviors, the preferred solution falls within the reinforcement learning (rl) paradigm \cite{sutton2018reinforcement}. Moreover, due to the very large (practically infinite) state and action spaces in real life problems, a promising approach heavily researched is based on the AlphaZero algorithm \cite{MasteringGoNature,AlphaGo2018Sceince} which is a model-based algorithm allowing us to perform an evaluation of the specific state encountered in real-time and use off-line learned insights only as prior information \cite{Volvo2019}.
The AlphaZero algorithm builds on the Monte Carlo Tree Search (MCTS) approach \cite{SurveyMCTS, recentModificationsMCTS}, which gradually builds a tree from the current state (which is the root of the tree), to depict the possible futures (the evaluation) resulting from different sequences of actions. 

The motivation for this work is the need to explain the decisions of the planning system to an engineer or user that is experiencing that planner's resulting behavior. As such, this work falls in under explainable AI (XAI) \cite{doshivelez2017rigorous,AdadiBerrada,RosenfeldAndRichardson,nguyen2020quantitative,Hoffman2018MetricsFE}. Explanations designed specifically for automotive applications, meaning during automated functions of the vehicle, were also considered in \cite{ZHOU20211}, \cite{Beggiato} and \cite{GoldmanBustin2022}. 
We are considering XAI for sequential decision making, which has also received some attention. 
A wider review of this problem that specifically focused on contrastive explanations was done in \cite{Magazzeni}.
Explanations as model reconciliation, meaning the explanations come in light of the differences between the model of the AI system and the user model were examined in \cite{Chakraborti,Sreedharan2019PlanningWE,SREEDHARAN2021103558}. Specifically, in \cite{Sreedharan2019PlanningWE} the suggested approach is to combine the extraction of explanations with the planning problem. 
The explanations considered in \cite{OfraAmir} are policy summarization. The goal of policy summarization is to convey the strengths and weaknesses of the agent by demonstrating their behavior in a subset of informative states. Another work that follows a similar line, but focuses on the explanability of RL agents is \cite{Sequeira_2020}.   
Another approach, considered in \cite{Kulkarni}, focuses on producing more explicable planning results. They do so by learning an explicability distance function between the AI plans and human's input. This learned explicability distance function is then used as a heuristic to guide the search towards explicable plans.
In \cite{lam2021identifying} the authors considered a setting similar to ours, with an RL agent that uses search trees to reach its decision. The problem they are considering is the ability of humans (AI experts) to identify reasoning flaws. They too are faced with the complexity of the search tree, and suggest the construction of a user interface to assist in the identification process. The work we suggest here could be a valuable tool to support such efforts. 
Finally, \cite{Baier2020TowardsEM}, considers the same problem of explainability of MCTS, and gives some high-level directions towards its solution. 


The XAI for sequential decision making in automated driving is a very complex problem. To simplify the problem we suggest that it can be studied by answering three main questions: \emph{When}, \emph{What} and \emph{How} to explain. The \emph{when} refers to the decision of whether or not to provide an explanation at any point in time during the ride. The \emph{what} refers to the actual content of the explanation, and the \emph{how} refers to the modality in which this explanation will be conveyed. These questions are not independent\footnote{For example, assuming there is a single dominant action, meaning the planner is very certain, this might lead us to provide an explanation, and also determine the type of explanation that we will provide, as an explanation focusing on the dominant choice, answering also the \emph{what} part.} 
and the split comes only as a method to simplify the overall problem. 
%
%
Our focus in this work is on the extraction of information from the planner that can assist in determining the above mentioned aspects of the explanations. We shortly provide concrete examples of usage of the methods suggested here relating to both the \emph{when} and \emph{what} questions. 


As mentioned, we consider the AlphaZero algorithm, and more specifically the real-time construction of the MCTS. As a stepping stone, we explore the question of XAI for sequential decision making in a simplified setting in which the action space is a discrete set of seven possible actions: \emph{acceleration regular, acceleration harsh, deceleration regular, deceleration harsh, right, left} and \emph{none} (where \emph{right} and \emph{left} refer to a change in the latitudinal location of the vehicle, a lane change for example, and \emph{none} refers to maintaining the same speed and lane).

The MCTS resulting data structure is built gradually. At each step of this process the algorithm chooses which additional node to add to the tree. These decisions comprise a significant part of the reasoning process of the algorithm, eventually leading to the resulting MCTS and to the chosen action the algorithm outputs for the current state (one of the actions from the root node of the MCTS). Consequently, the resulting MCTS's structure summarizes this gradual construction process and gives us a window to the sequence of choices of the algorithm - the reasoning of the algorithm. This central observation led us to consider the resulting MCTS data structure as a type of explanation. However, these trees have two distinctive weaknesses. The first is that these trees can become very big and very complex, usually with a lot of details that one would prefer to neglect in a concise explanation.
The second is that the information (including structure) is spread within all the nodes of the tree, requiring additional (sometime complex) computations to extract it. 

In this work we address both aspects: the size of the MCTS and the spread of the information across the tree. Our goal is to provide raw explanations, meaning explanations that are still in the form of a MCTS data structure, however more concise, meaning reduced in size, and with additional information that captures aspects of the reasoning process. Specifically, in this work we focus on information that attempts to capture the structure of the MCTS, being a window to its reasoning process. 




Given the above, we first tackle the issue of information relating to the structure of the MCTS. When we observe a tree with our eyes we can clearly determine its structure. We can state which actions the AI algorithm considered more, where it was confident in its search and where not. We suggest an approach to extract this structural information as numerical values relating, at each node, to the entire subtree spanned from it. Using this measure we then tackle the aspect of size. We examine a specific criterion for the reduction of the tree that attempts to reduce the overall size of the tree while minimizing the loss to information available in the tree. The suggested approach reduces the size of the tree specifically by removing complete subtrees\footnote{Another method for reducing the size of the tree is to unifying nodes.}. We provide some analytical properties of such reductions, emphasizing the complexity of the suggested reduction criterion, and suggest practical greedy algorithms that attempt to approximate the suggested criterion over the entire tree while updating the relevant numerical values within the tree. Finally, we provide concrete data regarding the performance of these reduction algorithms, comparing their performances in different aspects, such as the general number of nodes and depth in the reduced trees.  

These raw explanations, in the form of concise MCTS, are intermediate data structure in the path towards the construction of different human understandable explanations. As an example, given a MCTS with structure values at every node, one can decide to provide an explanation for the chosen action only when the structure value of the chosen action's subtree is very low, indicating that the algorithm was relatively certain of the next action to take at all future steps (thin structure). In this example the structure value is used to determine \emph{when} to provide an explanation. Alternatively, we can use the structure value for the \emph{what}, meaning, directly within an explanation. For example, by comparing the normalized value of all actions considered at the root node, an explanation could be for example “The algorithm is more confident in its future choices when following this action compared to all other possible actions” (when the structure value of the chosen action is lower than the alternative actions), or “The chosen action opens many possibilities for future actions” when the structure value of the chosen action is higher than the alternative actions. These are simple examples that do not require additional processing and thus can be performed on the original MCTS with the added structure values, however one can think of other more complex constructions of explanations. For these, the reduction becomes significant, as it reduces computations. As an example of such processing consider the comparison of two MCTS. Since a MCTS is constructed at every time step, spanning possible futures we expect the current MCTS to be similar to the subtree spanned from the chosen action at the previous time step. This similarity might, at times, be broken, indicating a deviation. By comparing these MCTS we can identify such deviations and provide an explanation, e.g., “we identify a deviation between what we thought would happen when taking the previous action, and what actually resulted”. These are just a few examples to explain how both the structure measure and the reduction can be used as immediate tools to enhance our ability to construct human understandable explanations.


\section{Entropy of MCTS as a Measure of Structure} \label{sec:treeEntropy}

\begin{figure}
\includegraphics[scale=0.03]{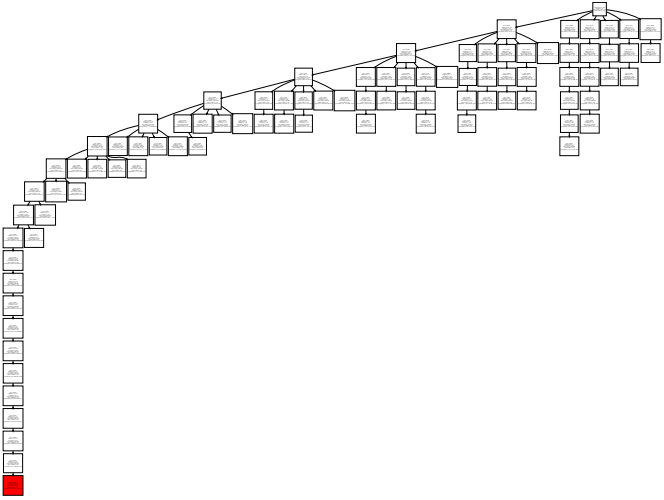}\caption{A MCTS example} \label{fig:MCTSexample} 
\hspace{1cm}
\end{figure}

The MCTS algorithm gradually builds a tree, starting from the current root state, by performing the following four steps over and over again until some time/computation budget is concluded \cite{SurveyMCTS}:
\begin{enumerate}
    \item Selection: starting from the root node, the selection process goes down the tree until it reaches a leaf node. The specific selection used is the Upper Confidence Bound applied to Trees (UCT) which balances exploitation of actions that have shown to be favorable, and exploration of actions that have not been sufficiently explored. 

    \item Expansion: when reaching a leaf node, it can be a terminal node, in which case it cannot be expanded. If the node is not a terminal node, the tree is expanded to the set of actions that can be considered from this node. 

    \item Simulation/Neural Network (NN): the properties of the expanded node can be evaluated using a Monte-Carlo simulation, where multiple roll-outs reaching a terminal state are performed to assess the value of the expanded node, and a prior distribution over its actions. In the AlphaZero family of algorithms the off-line learning of the value and prior per state (node) replace this.  

    \item Back-propagation: after expanding and extracting the value and prior distribution of the node, this new information is back-propagated all the way up the tree to the root node. The update is done for the accumulated value at each node and the number of visits at each node. 
    
\end{enumerate}
Thus, at every step the tree is expanded by a single node (excluding cases of terminal nodes in which there is no expansion). When the time/computation budget is over, the choice of action at the root node is usually determined as the most visited child of the root. An example of such a tree is given in Figure \ref{fig:MCTSexample}. At each node in the tree the vector corresponding to the number of visits of its children nodes can be considered a probability vector reflecting the choice of the algorithm to expand the tree in one direction over the other. This aspect, together with the gradual expansion are the main points used next. 

A tree, in which each node contains a probability vector, can be depicted as a discrete random process: 
\begin{align}
X_0, X_1, \ldots, X_n,
\end{align}
where $X_0$ is the random variable over the possible actions considered at the root node, distributed according to the probability vector available at the root. $X_1$ is the random variable over all possible actions from the children of the root node, where the probability can now be extracted from the probability at the root node and the probability vectors at each of its children. We continue this until we reach the leaf nodes of the tree. 

The entropy measure reflects the uncertainty of a random process. It is maximized if at each node all possible actions are considered and are equally likely (uniformly distributed). 
The more uniform the distribution the wider the tree, since all options were considered and expanded upon. On the other hand, the more deterministic the process is, meaning the probability is concentrated on less actions, and less actions were expanded further. In this case the tree is thinner and more concentrated. Thus, an entropy measure at a node gives us insight to the structure of the subtree expanded from that node. Moreover, having at our disposal the entropy measure at every node we enhance our ability to determine the structure of the tree at different levels of resolution. As explained above, the structure of the tree reflects important aspects in the reasoning of the algorithm and can be directly translated to an explanation. 
The entropy of trees with unbounded degree was also considered in \cite{EntropyTrees}, motivated by a different purpose of considering the lossless compression of these data structures.  

Our first goal is to provide efficient methods for calculating the entropy at each node, given that these trees are MCTS that have a specific construction process.
As this is a discrete random process, the standard method for calculating its entropy is as follows:
\begin{align} \label{eq:entropyRecursive_v1}
H \left( X_0, X_1, \ldots, X_n \right) & = H \left( X_0 \right) + H \left( X_1, \ldots, X_n | X_0 \right) \\
& = H \left( X_0 \right) + \sum_{i = 1}^n H \left(X_i | X_0, \ldots , X_{i-1} \right) \nonumber
\end{align}
where we used the chain rule. $H\left( X_0 \right)$ is easily calculated from the probability vector at the root. 
The conditional entropy requires more delicate care. To make this simpler we use the following notation:
\begin{align}
T_i \equiv X_0, \ldots, X_{i-1}.
\end{align}
This notation denotes the possible paths to a node at level $i$. $T_0$ is the empty path that leads to the root node of the MCTS. $T_1 = X_0$ and thus defines the paths that lead to the nodes that constitute the root nodes of the subtrees in depth one (children of the root node).
\begin{align}
H \left(X_i | X_0, \ldots , X_{i-1} \right) & = H \left(X_i | T_i \right) \\
& = \sum_{j \in \left[ | \mathcal{A} |^i \right]} \Pr( T_i = t_j ) H \left( X_i| T_i = t_j \right) \nonumber
\end{align}
where $\mathcal{A}$ is the action space, which determines the maximum number of children each node may have. The notation $\left[ x \right]$ denotes the set $\{0,1,\ldots, x-1 \}$. Thus, the maximum number of nodes at level $i$ is $| \mathcal{A} |^i$, assuming all actions are valid at each node.
Putting it together, we have that the entropy of a MCTS is:
\begin{multline} \label{eq:entropyMCTS}
H \left( X_0, X_1, \ldots, X_n \right) = H \left( X_0 \right) + \\ \sum_{i = 1}^n \sum_{j \in \left[ | \mathcal{A} |^i \right]} \Pr( T_i = t_j ) H \left( X_i | T_i = t_j \right).
\end{multline}
Equation \eqref{eq:entropyMCTS} suggests a recursive implementation, going down the tree from its root, and calculating the probabilities to each node throughout. 
The main issue with an implementation following \eqref{eq:entropyMCTS} is that since any expansion of the tree changes the probabilities (the probabilities from the root to each node) throughout the tree, such an expansion requires us to recalculate the entropy from scratch. 

As an alternative approach we go back to equation \eqref{eq:entropyRecursive_v1} and using the following notation:
\begin{align}
T^i \equiv X_i, X_{i+1}, \ldots, X_n
\end{align}
we can write it as follows:
\begin{align} \label{eq:entropyRecursive}
H \left( X_0, \ldots, X_n \right) = & H \left( X_0 \right) + H \left( X_1, \ldots, X_n | X_0 \right) \\
H \left( T^0 \right) = & H \left( X_0 \right) + H \left( T^1 | X_0 \right) \nonumber \\
= & H \left( X_0 \right) \nonumber \\ 
& + \sum_{j \in \left[ |\mathcal{A}| \right] } \Pr \left( X_0 = x_j \right) H \left( T^1 | X_0 = x_j \right). \nonumber
\end{align}
The above can be further extended since for each $j \in \left[ |\mathcal{A}| \right]$ we have that $H \left( T^1 | X_0 = x_j \right)$ is:
\begin{align}
& H \left( X_1 | X_0 = x_j \right) + H \left( X_2, \ldots, X_n | X_1, X_0 = x_j \right) = \nonumber \\
& H \left( X_1 | X_0 = x_j \right) + \nonumber \\ 
& \sum_{k \in \left[ |\mathcal{A}| \right] } \Pr \left( X_1 = x_k | X_0 = x_j \right) H \left( T^2 | X_1 = x_k, X_0 = x_j \right). \nonumber
\end{align}
We can clearly see that this writing suggests a different, and perhaps cleaner, recursive implementation. We can first observe that this recursive implementation builds, at every stage, on the calculation of the entropy of the subtrees - $H \left( T^1 | X_0 = x_j \right)$. The advantage of this is that we can produce directly from this formalization the entropy measure at each node of the tree, and not only at the root of the tree. A second significant difference is that the multiplications are with conditional probabilities. This difference is very significant since it reduces the computational impact as a result of changes done to the tree (such as expansions or reductions). 
The above expression can be written in a generalized manner as follows:
\begin{align} \label{eq:entropyMCTS_recursionGeneral}
& H \left( T^i | T_i = t \right) = H \left( X_i | T_i = t \right) + \\
& \sum_{j \in \left[ |\mathcal{A}| \right] } \Pr( X_i = x_j | T_i = t) H \left( T^{i+1} | T_i = t, X_i = x_j \right) \nonumber
\end{align}
and we can observe two elements. The first being $H \left( X_i | T_i = t \right)$ which is the entropy of the random variable of level $i$ given that $T_i = t$. The conditioning reduces this entropy to the entropy of a specific node at level $i$, determined by $T_i = t$. The summation element is a weighted summation of the entropy of the subtrees, where the weighing is according to the probability of each subtree.

The basic implementation following \eqref{eq:entropyMCTS_recursionGeneral} is to recursively go down the MCTS through every node. 
This approach is simple, clear, and works for any finalized tree. The main advantage is that it produces the entropy computation for every subtree as a by-product. More details are available in Appendix \ref{appendix:basicImplementationOfEntropy}. 

We focus on an alternative approach showing that \eqref{eq:entropyMCTS_recursionGeneral} can be seamlessly integrated into the MCTS construction. 
As mentioned, the trees constructed during the MCTS algorithm are gradually expanded by performing the four steps of: selection, expansion, simulation and back-propagation over and over again.
A more efficient calculation of the entropy is to integrate the calculation into the construction of the MCTS. This means that we constantly update the values of the entropy as the MCTS evolves, and those are available anytime during the construction. 


Similar updates already occur in the MCTS algorithm. The value and number of visits are updated up the tree during the back-propagation step of the MCTS algorithm, going from a leaf all the way to the root. Similarly, the entropy can be updated. Thus, at every iteration of the MCTS algorithm the entropy value is updated and fits the current number of visits values of the tree, meaning it fits the tree of that stage. 
Thus, we essentially change the back-propagation step of the MCTS algorithm as shown in Algorithm 
\ref{algo:MCTSentropy_backpropagation} where the actual entropy update step is given in Algorithm 2\footnote{Lines \ref{algo:lineStartingFor} - \ref{algo:lineEndingFor} are written as a for-loop for clarity. These computations can be vectorized.}. 

\begin{algorithm}[t]
\caption{new MCTS back-propagation with entropy}\label{algo:MCTSentropy_backpropagation}
\begin{algorithmic}[1]
\Procedure{BackPropagation}{$leaf, value, root$}
\State \% \emph{For leaf:}
\State $currentNode = leaf$
\State $currentNode.value = value$
\State $currentNode.numberVisits += 1$
\State $trajectoryDepth = 0$
\While {$currentNode$ is not $root$}
\State $value = currentNode.reward + \gamma \times value$
\State $currentNode = currentNode.parent$
\State $currentNode.value = value$
\State $currentNode.numberVisits += 1$
\State $trajectoryDepth += 1$
\State $currentNode.depth = $
\State $\quad \max( currentNode.depth, trajectoryDepth)$
\State updateEntropy($currentNode$)
\EndWhile
\EndProcedure
\end{algorithmic}
\end{algorithm}

\begin{algorithm}[t]
\caption{Entropy calculation during back-propagation}
\begin{algorithmic}[1] \label{algo:updateEntropy}
\Procedure{updateEntropy}{$node$}
\State \% \emph{$treePolicy$ (prob. vector over actions) calculation:}
\State $\mathbf{treePolicy} = $
\State $node.\mathbf{childNumberVisits} / node.numberVisits$
\If{$sum(treePolicy)$}:
\State $\mathbf{treePolicy} = \mathbf{treePolicy} / max(\mathbf{treePolicy})$
\State $\mathbf{treePolicy} = \mathbf{treePolicy} / sum(\mathbf{treePolicy})$
\EndIf
\State \% \emph{Initialization:}
\State $H_x = 0$
\State $H_{children} = 0$
\State \% \emph{Computation of entropy:}
\For{$index$ $1$ to length($\mathbf{treePolicy}$)} \label{algo:lineStartingFor}
\State $probability = \mathbf{treePolicy}[index]$
\If{$probability$}
\State $H_x += -probability * \log_2(probability)$
\If{$index$ in $node.children$}
\State $H_{children} += probability * $
\State $\quad node.children[index].entropy$ \label{algo:lineEndingFor}
\EndIf 
\EndIf
\EndFor
\State $node.entropy = H_x + H_{children}$
\EndProcedure
\end{algorithmic}
\end{algorithm}



The calculated entropy of the tree is the entropy of the process, starting from the time point of the root node all the way to the leaf. At every node we calculate the entropy, and for that calculation, that node takes the role of the root. This means that the random processes for which we are calculating the entropy are of different lengths. For example, consider siblings of a specific node. The entropy in each of these siblings is of a random process of a different length, depending of the depth of the subtree expanded from each sibling. This makes it impossible to compare between the two entropies. The longest process is always the one starting at the root of the tree. Going down the tree processes get shorter and shorter, and thus also their entropies decrease. 

If we want to compare different values of entropy in different nodes in the tree we need to normalize their values and produce an estimate of the entropy per time-step (per level of the tree). This is again problematic, as it is not clear how to normalize the entropy value at a node where one of the children has a very long and deep subtree while the other child has a short (shallow) subtree.
Thus, we try to estimate the entropy per time-step using lower and upper bounds. For this purpose we also back-propagate the depth value, as can be seen in Algorithm \ref{algo:MCTSentropy_backpropagation}. 

Assume $N$ denotes the number of nodes in a given subtree.
The tree could be even and each level full. In such a case, the connection between $N$ and ${\ell}$, the depth of the tree, is given by the following equation:
\begin{align} \label{eq:depthNumNodes}
N = \sum_{i=0}^{{\ell}} |\mathcal{A}|^{i} = \frac{ |\mathcal{A}|^{{\ell}} - 1}{|\mathcal{A}| - 1}
\end{align}
where $\mathcal{A}$ is the action space (maximum number of children of each node) and assuming $|\mathcal{A}| > 1$. However, this is usually not the case. Thus, given $N$ we can lower bound $\ell$ as follows:
\begin{align} \label{eq:LBonDepth}
\ell \geq \log_{|\mathcal{A}|} \left( N ( |\mathcal{A}| - 1) + 1 \right) \equiv \ell_{LB}.
\end{align}
Note that the above bound can be sharpened if we keep track of the number of children the node expanded, and back propagate the maximum value, thus replacing $\mathcal{A}$ with a smaller value, $\mathcal{A}_{\textrm{in use}}$, in some cases. This is important specifically in cases where the action space is large, but usually only a small subset of it is begin explored (or even valid).
%
%
Denoting the average, per time-step, entropy as $H( \bar{X} )$ we thus have that
\begin{align}
\frac{H\left(X_0, X_1, \ldots, X_{\ell} \right)}{\ell} \leq H( \bar{X} ) \leq \frac{H\left(X_0, X_1, \ldots, X_{\ell} \right)}{\ell_{LB}}.
\end{align}

In two extreme cases the above lower and upper bound equal. The first is the case were the inequality in \eqref{eq:LBonDepth} holds with equality. This is the case when the subtree is even (over the entire set of actions $\mathcal{A}$, or the smaller set $\mathcal{A}_{\textrm{in use}}$). The second case is when we have a single trajectory. In this case equation \eqref{eq:LBonDepth} is not valid since |$\mathcal{A}_{\textrm{in use}}| = 1$ and the depth, $\ell$, equals both the number of nodes $N$ and $\ell_{LB}$ (calculated directly using equation \eqref{eq:depthNumNodes}). Thus, again the lower bound and upper bound equal.

In practice, and as shown in Algorithm \ref{algo:MCTSentropy_backpropagation} the depth of each node is calculated and $N$ is provided from the $numberVisits$ variable. The calculation of the $|\mathcal{A}_{\textrm{in use}}|$ at each node for, reflecting the maximum number of actions examined in the subtree expanding from this node, can also be added to sharpen the bounds, throughout.



\section{Reduction of MCTS using Subtree Removal} \label{sec:removal}

So far we tackled the extraction of information spread within the tree to numerical values that capture the structure, and hence the reasoning of the AlphaZero algorithm. We have shown that this can be done efficiently. Another usage of these numerical entropy values is in tackling the aspect of size. 

Reduction of a tree can be done in several ways. One can unify two nodes into one, or summarize two subtrees into a single subtree. In this work we specifically consider the reduction of a tree using only subtree removals. This means, that the tree can be reduced by picking the set of nodes that will be removed, where by removing a node we remove the entire subtree spanned from it. 

Any form of removal usually has some guiding criteria defining when to perform a removal. The specific choice of the criterion depends on the intended usage of the reduced tree. Our goal is to obtain a smaller, and more concise MCTS. We want to reduce the size of the MCTS, while still holding to the information it conveys. Such information contains, for example, the regions that the algorithm expanded more, the main children that were expanded at each level, etc. Thus, we propose a criterion that examines this exact trade-off. On the one hand we wish to reduce the size of the MCTS, meaning the number of nodes, while on the other hand we have the entropy of the MCTS which captures the interest. The trade-off between entropy and size should leave us with a MCTS that is smaller and focused on complex areas that are of interest. In mathematical terms this can be written as:
\begin{align} \label{eq:optimizationTradeOff}
\max \left\{ H \left( T^0 \right) - \beta \log \left( |\mathcal{A} | \right) numberOfNodes \right\} .
\end{align}
where $\beta$ is the trade-off factor, $numberOfNodes$ can be approximated by the value of the $numberOfVists$ in MCTS (as most visits result with an expansion of the tree, meaning they increase the tree by one node) and $\mathcal{A}$ is our action space. 

In order to make the two terms comparable we consider both terms as descriptive information. The first term, the entropy, provides us with the number of bits\footnote{We assume $\log$ to the base of 2, throughout.} required to describe the tree given the probability vectors that were extracted from the number of visits. For the second term, we consider a worst case tree with $numberOfNodes$ nodes, in the sense of description. An upper bound on the longest description is attained by a uniformly distributed tree over all possible actions, $|\mathcal{A} |$, at each node. 
The range of relevant values to consider for $\beta$ are $[0, \beta_{ub}]$ where:
\begin{align} \label{eq:beta_UB}
\beta_{ub} \equiv \frac{ H \left( T^0 \right) }{\log \left( |\mathcal{A} | \right) numberOfNodes}.
\end{align}



Examining equation \eqref{eq:optimizationTradeOff} we have a clear understanding of how the right element behaves when we remove a subtree, as it is monotonically decreasing in $numberOfNodes$. $H \left( T^0 \right)$, on the other hand, is a bit more complex. This is the entropy of the entire tree. When we remove a subtree somewhere within the tree, the entropy of the entire tree changes. However, it does not necessarily increase or decrease. The next theorem shows exactly how the entropy of a node changes, given that one of its immediate children has been removed. 

\begin{theorem} \label{thm:main}
Assume a node at depth $i$ defined by $T_i = t$, with $N$ number of visits accumulated at its children. Its entropy is given by equation \eqref{eq:entropyMCTS_recursionGeneral}. 
When we remove child $k$ the entropy of this node is as follows:
\begin{align} \label{eq:theorem1exp}
& H \left( \tilde{T^i} | T_i = t \right) \nonumber \\
& = H \left( T^i | T_i = t \right) + \frac{1}{1- \Pr( X_i = x_k | T_i = t)} \cdot \nonumber \\
& \Bigl( \Pr( X_i = x_k | T_i = t) H \left( T^i | T_i = t \right) -H_b(\Pr ( X_i = x_k | T_i = t)) \Bigr. \nonumber \\ & \Bigl. - H \left( T^{i+1} | T_i = t, X_i = x_k \right) \Bigr), 
\end{align}
where $H_b( \cdot )$ is the binary entropy.
\end{theorem}
\begin{proof}
    Proof is available in Appendix \ref{appendix:proofTheorem1}.
\end{proof}

Theorem \ref{thm:main} gives us the new entropy of the node whose child has been removed in terms of the original entropy of that node (before the removal) - $H \left( T^i | T_i = t \right)$. We can see that not in all cases will the entropy decrease when a child node is removed. Whether a removal will decrease or increase the entropy, depends on the expression in the last two rows of \eqref{eq:theorem1exp}. 
These rows examine the difference between a fraction of the original entropy (according to the probability of the removed child) and the contribution of the removed node given in two terms: the binary entropy of the probability of the removed child and the entropy contribution of the removed child. This is to be expected, and gives us an exact expression to examine whether a specific removal will increase or decrease the entropy of its parent node.   

Theorem \ref{thm:main} considers the effect of a subtree removal on the entropy of the immediate parent node, however, we are more interested in how a removal effects the entropy of the entire tree. Recall that a removal of a subtree changes not only the probability vector of the parent node (and hence its local entropy contribution), but also the probability vectors of all of its ancestors. 

The next theorem considers the more general effect by considering the change to the entropy of a parent node given that the entropy of one of its children has changed. This result can be used over and over to propagate the change in entropy up the tree. 

\begin{theorem} \label{thm:general}
Assume a parent node at depth $i$ defined by $T_i = t$, with $N_p$ number of visits accumulated at its children (counts). Further assume that the $\ell^{th}$ child of this node had a change in its subtree that resulted with a decrease in the accumulated number of visits of its children from $N$ to $N (1 - \tilde{p})$ and as a result of this change the entropy of this child changed from 
$H_{\ell} \equiv H \left( {T}^{i+1} | T_i = t, X_i = x_{\ell} \right)$
 to 
$\tilde{H}_{\ell} \equiv H \left( \tilde{T}^{i+1} | T_i = t, X_i = x_{\ell} \right)$.
The entropy of the parent node is as follows:
    \begin{align} \label{eq:generalChangeInEntropy}
        & H \left( \tilde{T}^i | T_i = t \right) =  \\
        & H \left( T^i | T_i = t \right) + \frac{1}{1- \hat{p}} \Bigl( \hat{p} H \left( T^i | T_i = t \right) -H_b(\hat{p}) \Bigr. \nonumber \\  & \Bigl. +(\Pr ( X_i = x_{\ell} | T_i = t) - \hat{p}) \cdot \Bigr. \nonumber \\
        & \Bigl. \left( H \left( \tilde{T}^{i+1} | T_i = t, X_i = x_{\ell} \right) - H \left( T^{i+1} | T_i = t, X_i = x_{\ell} \right) \right) \Bigr) \nonumber
    \end{align}
where 
\begin{align} \label{eq:p_hat}
    \hat{p} = \tilde{p} \left( \Pr ( X_i = x_{\ell} | T_i = t) - \frac{1}{N_p} \right). 
\end{align}
\end{theorem}
\begin{proof}
    Proof is available in Appendix \ref{appendix:proofTheorem2}.
\end{proof}

Note that Theorem \ref{thm:general} is not a generalization of Theorem \ref{thm:main}. Theorem \ref{thm:general} allows us to concatenate the change up the tree, while Theorem \ref{thm:main} considers a local removal of a subtree. Thus, in practice, we will first use Theorem \ref{thm:main} to assess the change of the entropy after the local removal and then we will iteratively use Theorem \ref{thm:general} to propagate the change up the tree. 

From the above expression we clearly see that in order to evaluate the effect of a removal on the entropy of the tree we need to examine only the trajectory path from the changed node all the way to the root. At each node we need only the original entropy value, the counts and the probability of the changed node.

The proof of the theorems requires the following lemma and more specifically the corollary that follows\footnote{We assume both are known but could not find the exact formalism, so we provide the proof in Appendix \ref{appendix:proofLemma}.}:
\begin{lemma} \label{lem:generalChangeLocalEntropy}
Assume a probability vector:
\begin{align} \label{eq:probabilityVectorP}
    \bold{p} = \bigl(p_1, p_2, \ldots, 1- \sum_{i} p_i \bigr).
\end{align}
Now assume that we change it to the following vector:
\begin{align}
    \bold{\tilde{p}} = \bigl(\frac{p_1}{1-\hat{p}}, \frac{p_2}{1-\hat{p}}, \ldots, \frac{p_{\ell-1}}{1-\hat{p}}, \frac{p_{\ell} - \hat{p}}{1-\hat{p}}, \frac{p_{\ell+1}}{1-\hat{p}}, \ldots, \frac{1- \sum_{i} p_i}{1-\hat{p}} \bigr)
\end{align}
where $\hat{p}$ is some probability value in $(0,p_{\ell})$ (note that the change to the $\ell^{th}$ component is different).  
The change to the entropy is as follows:
\begin{align}
    H(\bold{\tilde{p}} ) = \frac{1}{1-\hat{p}} H(\bold{p}) - \frac{1}{1-\hat{p}} H_b(\hat{p})
\end{align}
where $H_b( \cdot )$ is the binary entropy.
\end{lemma}

\begin{proof}
    Proof is available in Appendix \ref{appendix:proofLemma}.
\end{proof}

\begin{corollary} \label{cor:changeLocalEntropy}
Assume a natural number $N > 1$ and a probability vector $\bold{p}$ with entropy denoted as $H(\bold{p})$. The following is the vector of appearances:
\begin{align} \label{eq:originalProbability}
    N \bigl(p_1, p_2, \ldots, 1- \sum_{i} p_i \bigr) \equiv N \bold{p}.
\end{align}
By removing the $\ell^{th}$ element from this vector, meaning that we consider the following probability vector:
\begin{align} \label{eq:probabilityAfterRemoval}
\bold{\tilde{p}} = \left( \frac{Np_1}{\tilde{N}}, \frac{Np_2}{\tilde{N}} \ldots \frac{Np_{\ell-1}}{\tilde{N}}, \frac{Np_{\ell+1}}{\tilde{N}}, \ldots \frac{N}{\tilde{N}}(1 - \sum_i p_i) \right) 
\end{align}
where $\tilde{N} = N - Np_{\ell}$, the entropy changes according to:
\begin{align}
H(\bold{\tilde{p}} ) & = 
\frac{1}{1-p_{\ell}} H(\bold{p}) - \frac{1}{1-p_{\ell}} H_b(p_{\ell}) \nonumber \\
& = H(\bold{p}) + \frac{1}{1-p_{\ell}} \left[ p_{\ell} H(\bold{p}) - H_b(p_{\ell}) \right] 
\end{align}
\end{corollary}

The proofs of both theorems and Lemma \ref{lem:generalChangeLocalEntropy} are given in the supplementary material.


Knowing how the entropy changes due to a removal does not resolve the suggested optimization given in \eqref{eq:optimizationTradeOff} as we are examining a trade-off between size and entropy. Surely, a removal that increases the overall entropy is very good for us in terms of the trade-off, as it both increases the entropy and reduces the overall size. However, subtree removals that decrease the entropy of the overall tree are also potentially good removals, since they might still increase the overall trade-off. 

Moreover, Theorem \ref{thm:main} and \ref{thm:general} suggest that the complexity of evaluating the impact of a reduction or performing a reduction is proportional to the depth of the reduction.


\section{Implementation \& Empirical Evaluations} \label{sec:algorithmsAndExperiments}

\begin{figure}[h]
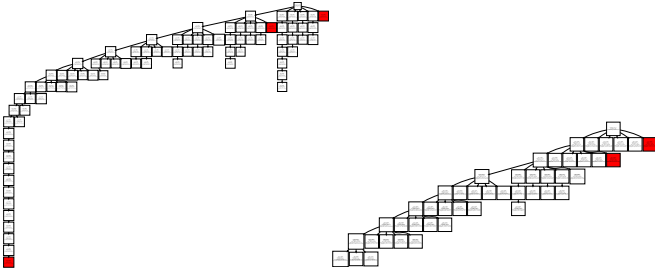

\centering
\begin{subfigure}[t]{0.5\linewidth}
    \centering
    \includegraphics[width=\linewidth]{figures/curve_2staeV2_halfBeta_original_180.pdf}
\end{subfigure}\hfil
\begin{subfigure}[t]{0.5\linewidth}
    \centering
    \includegraphics[width=\linewidth]{figures/curve_2staeV2_halfBeta_reducedTradeOff_180_summarized.pdf}
 \end{subfigure}
 \caption{An example of the reduction using the two-stage V2 algorithm. 
 $\beta = \frac{1}{2} \beta_{UB}$ (curve scenario). \newline}
 \label{fig:example_curve_two_stageV2_180}
\hspace{3cm}
\end{figure}

\begin{figure}[h]
\centering
\begin{subfigure}[t]{0.5\linewidth}
    \centering
    \includegraphics[width=\linewidth]{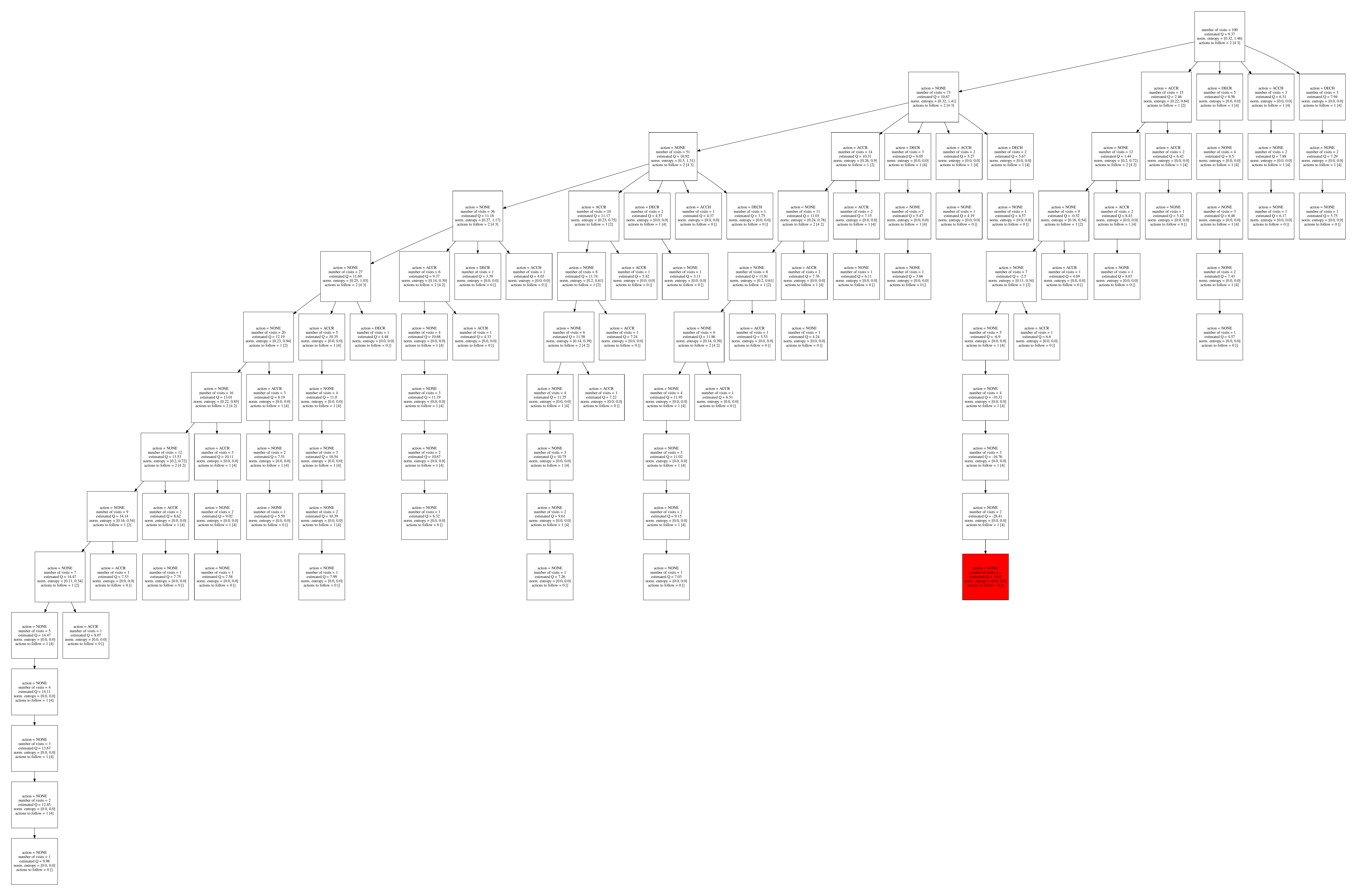}
\end{subfigure}\hfil
\begin{subfigure}[t]{0.5\linewidth}
    \centering
    \includegraphics[width=\linewidth]{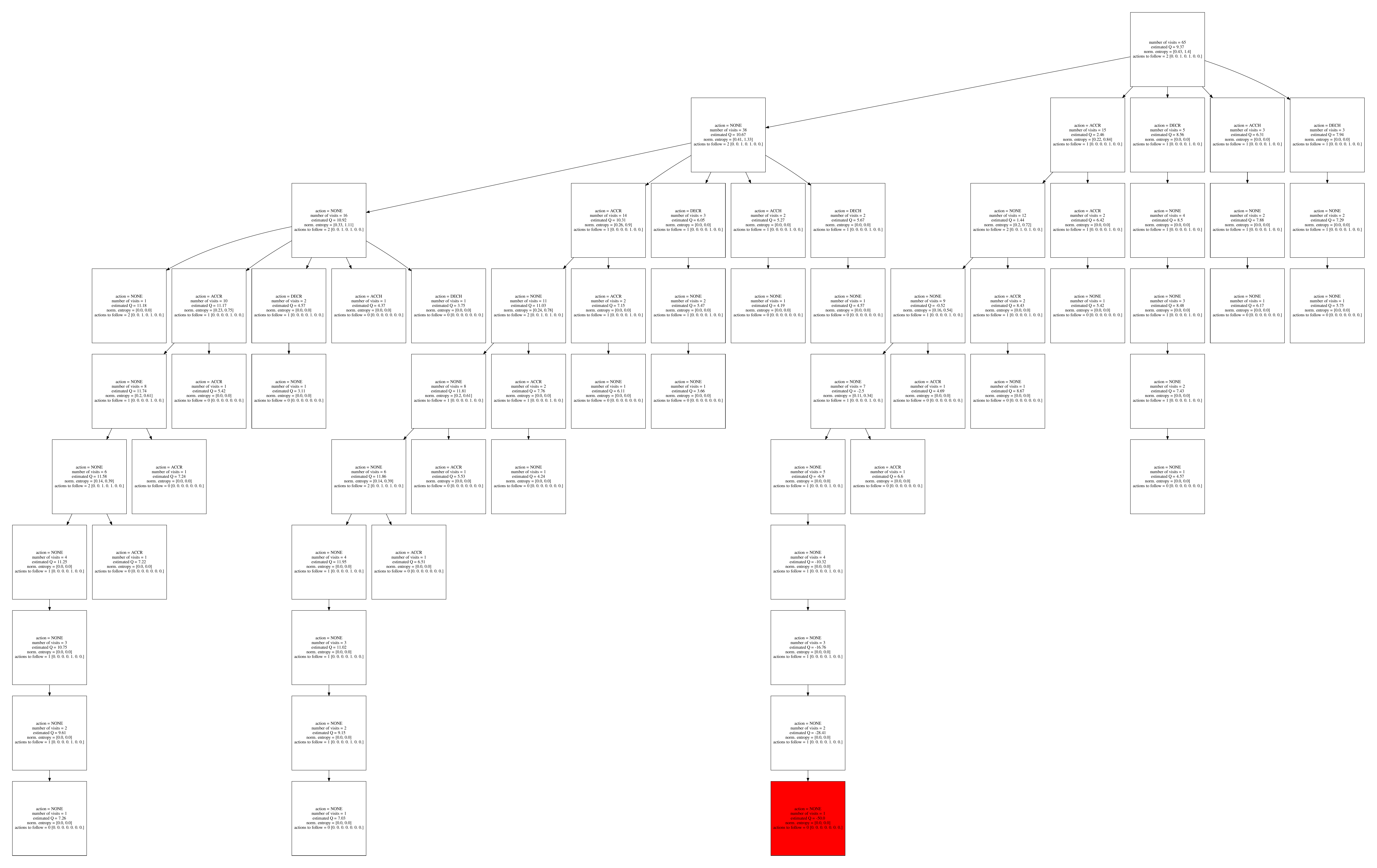}
 \end{subfigure}
 \caption{An example of the reduction using the two-stage with criterion (second variant) algorithm. 
 $\beta = \frac{1}{2} \beta_{UB}$ (merge scenario).\newline}
 \label{fig:example_merge_two_stage_with_criterion_50}
\hspace{3cm}
\end{figure}

There is a gap between our theoretical ambition to achieve a reduction according to the entropy vs. size trade-off and its possible feasible implementations. The two central issues are, first, that there are $2^{numberOfNode}$ possible subsets of reductions to consider. Second, the benefit of a specific subtree reduction might change once another subtree reduction is performed. Putting these two together makes the complexity of finding the optimal solution unaffordable. Thus, we chose greedy methods to perform reduction that are motivated by the trade-off given in equation \eqref{eq:optimizationTradeOff}. The first is a local approach that goes down the tree and for each node finds the optimal reduction set, meaning the set of children of that node that if removed would produce the highest trade-off, and performs their reduction. The second base-line approach is a two-stage approach. It goes down the tree and finds the best possible reduction set from each node. This reduction is \textbf{not} performed but rather placed in a priority list - $pList$. This priority list is kept sorted from best reduction set to worst. Note that the trade-off values determining the performance of a specific reduction are measured assuming that this it the only reduction performed on the entire tree\footnote{This is true for most nodes apart from cases when one reduction is an ancestor of another. See supplementary for more details}. The second stage is to go over $pList$. We consider three variants of the second stage. The first variant performs all reductions in $pList$. The second variant performed the reductions in $pList$ until we reach a reduction that no longer improves the trade-off of the entire tree in its new form (after reductions performed thus far). The last variant, denoted as two-stage V2, examines each element in $pList$ whether the specific reduction set improves the trade-off of the entire tree in its new form. The difference between the second and last variants is that the last variant continues to go over $pList$ even if a bad element, which did not improve the trade-off has been encountered. 
The one-stage approach gives more preferences to reductions up the tree, since those are encountered first and performed. It has the advantage that future potential reductions are evaluated on the already reduced tree. Alternatively, the two-stage approach evaluates each node as if no reductions have been performed, and by that allows all potential reductions to compete fairly. This emphasizes the difference between the one-stage approach and the first variant, which although performing all reductions in the priority list, these can be very different reductions than those performed by the one-stage approach.   
The complexity of both base-line algorithms is $O(N 2^{|\mathcal{A}|} \log N)$, where $N$ is the number of nodes in the tree, and $2^{|\mathcal{A}|}$ is included to show the dependency on the cardinality of $\mathcal{A}$. We provide here two examples of tree reductions in Figures \ref{fig:example_curve_two_stageV2_180} and \ref{fig:example_merge_two_stage_with_criterion_50}. Additional examples and algorithmic details are available in Appendix \ref{appendix:implementation_reduction}.

The performances of all four algorithms were examined on two simulated driving scenarios, a curved highway, and a ramp-merge. 
We used the open source traffic simulator SUMO \cite{SUMO2018} and the flow framework \cite{Wu2017FlowAA} for both training and evaluation of AlphaZero with the limited action space of seven actions. A decision step was performed every 1sec of simulation time. We used a discounting factor of $\gamma = 0.9$, identical weights for the exploration and exploitation elements in the Upper Confidence Tree (UCT) and a computation budget of $60$ iterations for the construction of the MCTS during training and $100$ during evaluation (meaning that when no terminal states are encountered the resulting evaluation trees contain $100$ nodes). Evaluations were done after limited training (10-15 iterations). For each of the four algorithmic settings, we evaluated each scenario by performing five evaluation drives, each such evaluation drive resulted with $25-80$ trees. 

The suggested reduction algorithms, which we applied on every single tree obtained, receive a criterion as input. 
The suggested criterion, given in \eqref{eq:optimizationTradeOff}, depends on the trade-off parameter $\beta$. We evaluated each of the four algorithms for three different values of $\beta$: $\beta_{UB}, \frac{1}{2}\beta_{UB}$ and $\frac{1}{4}\beta_{UB}$, where $\beta_{UB}$ is given in \eqref{eq:beta_UB} and depends on the specific tree considered. 

After applying the four reduction algorithms on each MCTS, for all three values of $\beta$, we compared the reduced versions with the original version. We examined several aspects including the change in entropy, size and trade-off between the original MCTS and its reduced form. We also examined the effect of the main path (path of highest probability down the tree), and main subtree (subtree of highest probability action from root), and similarly the effect on the secondary path and subtree (where secondary is the highest probability path and subtree starting from the second highest probability action from the root). All numerical results are provided in Appendix \ref{appendix:ExperimentalResults}. Here we provide the graphs examining the trade-off, entropy and size only for the ramp-merge scenario. The change in these three values was measured as the difference between the original value and the new value divided by the original value. Thus, when the result is negative it means an increase, whereas a positive result is a decrease. We expect the trade-off to increase (negative result) and the size to decrease, however the entropy can go either way, and we hoped to see an increase. 

\begin{figure}
\begin{center}
    \includegraphics[scale=0.4]{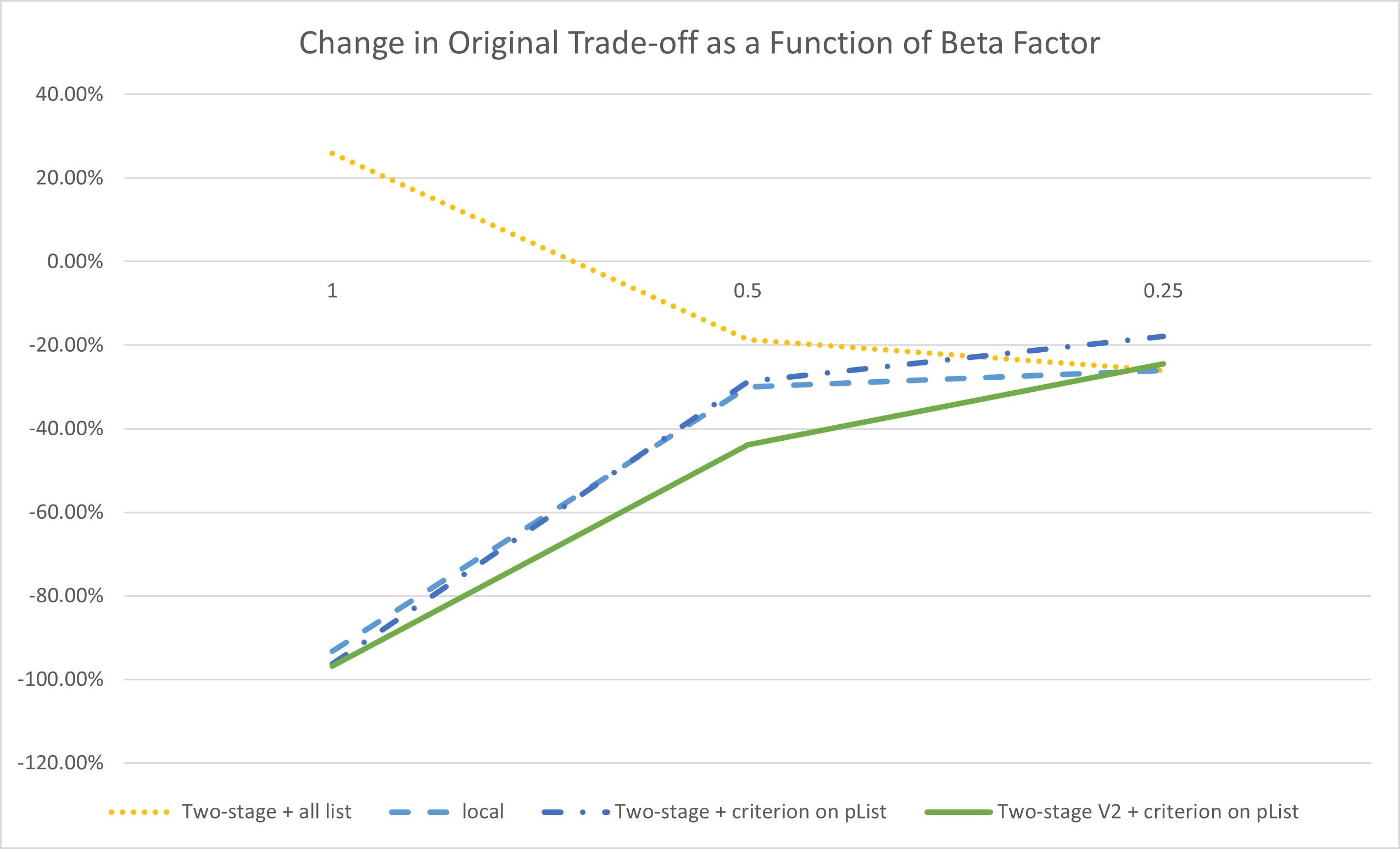}\caption{Comparing the change in trade-off in all three algorithms as a function of the factor of $\beta_{UB}$ - curve \newline} \label{fig:curve_tradeoff} 
\end{center}
\end{figure}

\begin{figure}
\centering
\begin{subfigure}[t]{0.5\linewidth}
    \centering
    \includegraphics[width=\linewidth]{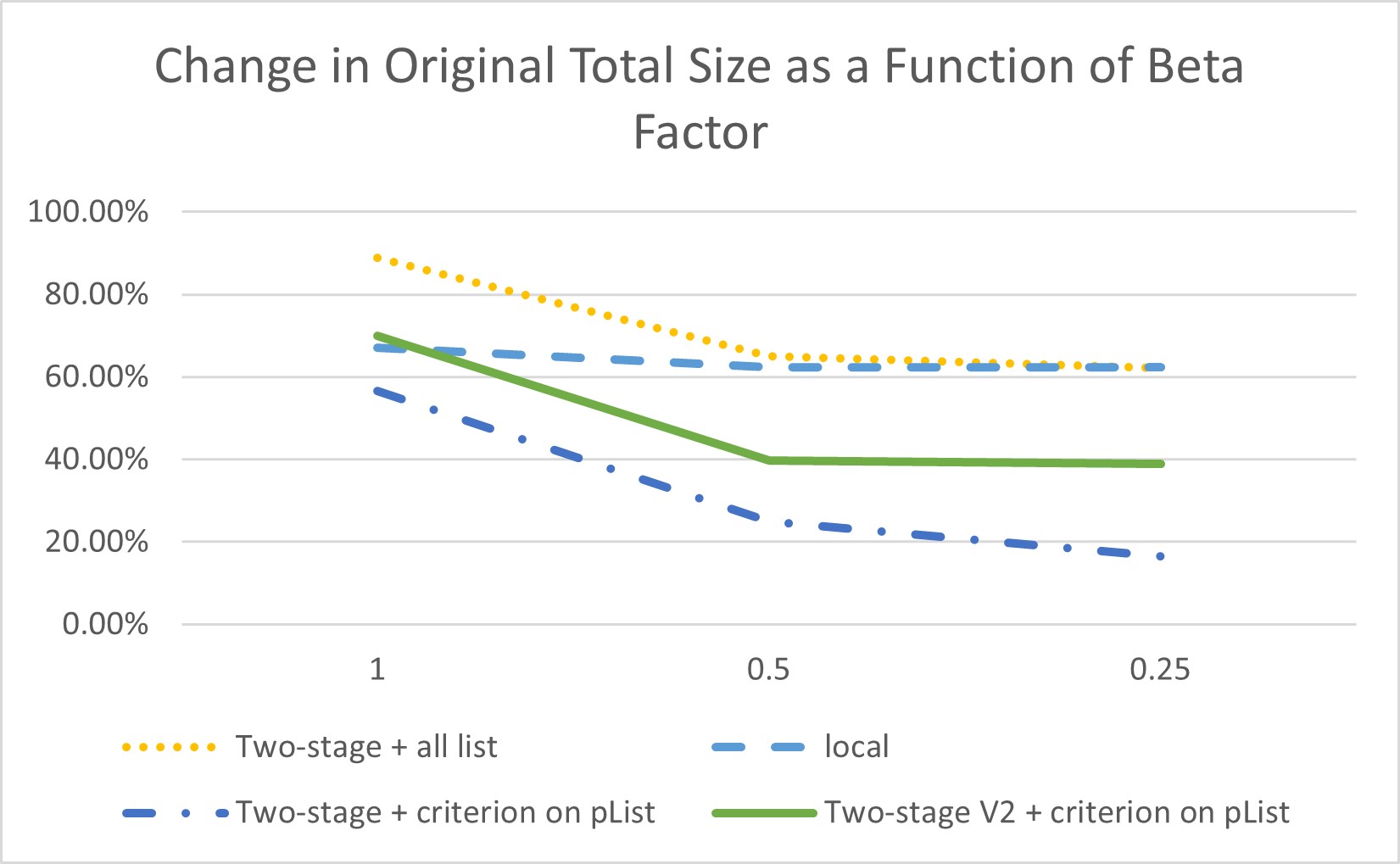}
\end{subfigure}\hfil
\begin{subfigure}[t]{0.5\linewidth}
    \centering
    \includegraphics[width=\linewidth]{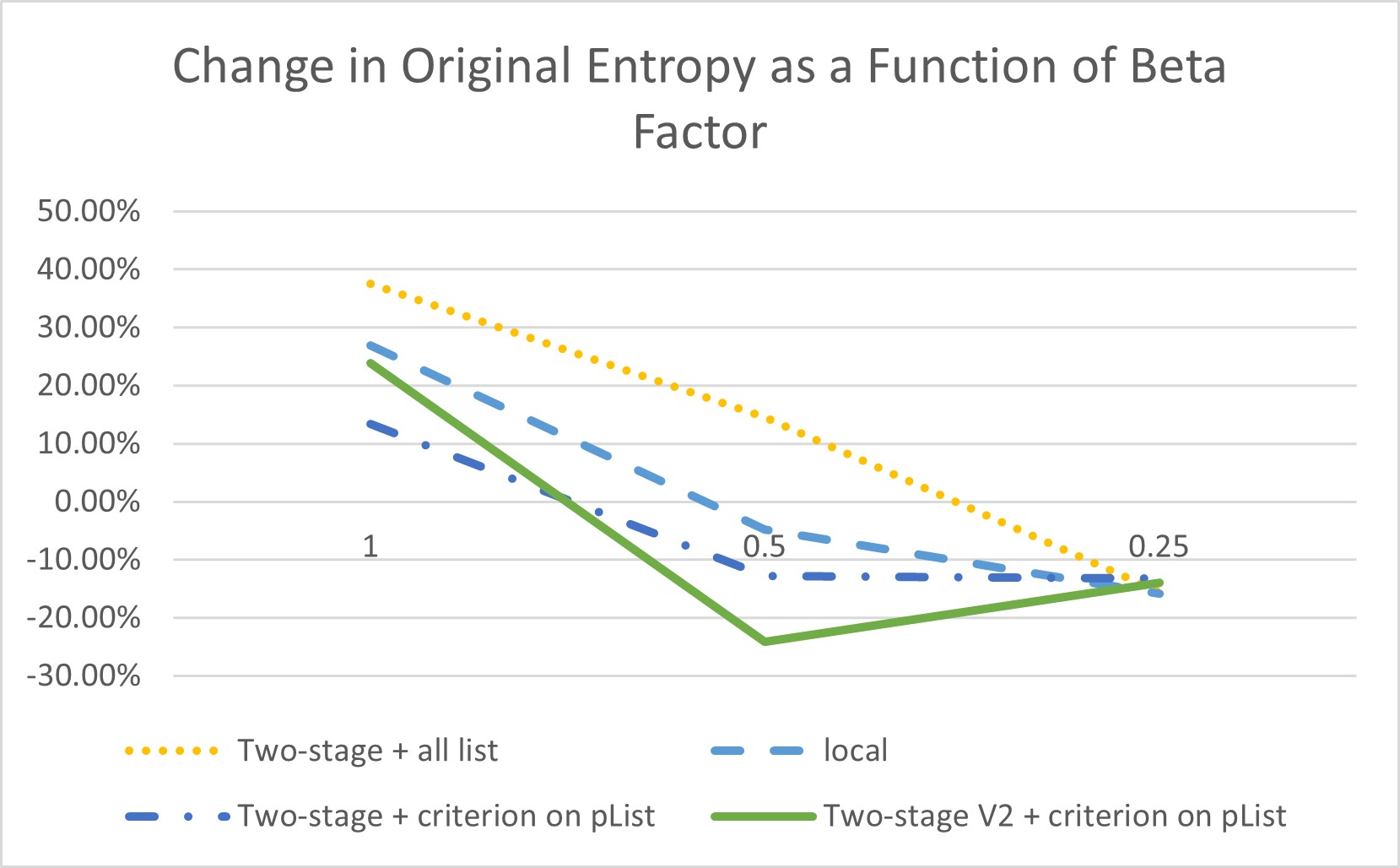}
 \end{subfigure}
 \caption{Comparing the change in total size and entropy in all four algorithms as a function of the factor of $\beta_{UB}$ - curve\newline} \label{fig:size_and_entropy_curve}
 \hspace{2cm}
\end{figure}

Figures \ref{fig:curve_tradeoff} and \ref{fig:size_and_entropy_curve} summarize the results on the change in trade-off, entropy and size as a function of the factor of $\beta_{UB}$ \eqref{eq:beta_UB} in the curve scenario. 
Similar graphs for the ramp-merge scenario are available in Appendix \ref{appendix:implementation_reduction}. 
%
%
We observe slight differences between the two scenarios of ramp-merge and curve. In both cases the best increase in trade-off was obtained for $\beta_{UB}$ (factor of 1), but it was obtained mostly by reducing the size, whereas the entropy reduced significantly ($30\%$ best case). Only for $\frac{1}{2} \beta_{UB}$ (factor of $0.5$) and lower do we see that some variants are able to preserve the entropy and even increase it slightly. 
Overall we see an advantage to the two-stage with criterion approach over a local or two-stage with no criterion approach (first variant). 
Specifically, the third variant of the two-stage approach (V2) is able to push towards further reductions while continuing to increase the overall entropy. 
Analysis related to the changes in depth of the main and secondary paths, as well as main and secondary subtree sizes can be found in Appendix \ref{appendix:ExperimentalResults}. 

\section{Summary} \label{sec:Summary}
In this work we tackle two aspects of the MCTS that are fundamental in our ability to support explainability of sequential decision-making algorithms that construct MCTS, such as AlphaZero. The first aspect is the spread of information throughout the MCTS, and specifically information related to the reasoning process of the algorithm. We show that this information, which influences the resulting structure of the MCTS, can be showcased as numerical entropy values that can be calculated efficiently during the expansion of the MCTS. The second aspect is the size of the resulting MCTS. We propose to use the entropy values as guiding values in the reduction of the tree in a trade-off criterion, thus preserve interesting attributes of the MCTS while attempting to reduce its size. We examine greedy algorithms motivated by the suggest criterion and evaluate their performance on MCTSs produced in driving simulations. 

These methods come as a tool to assist in the extraction of meaningful understandable explanations of the sequential decision-making algorithm. We have only began to apply them in such constructions, and see numerous different paths for the future.  

\appendix
\section*{Appendix}    

\section{Basic Implementation of Entropy Calculation} \label{appendix:basicImplementationOfEntropy}
A basic implementation of the calculation of the entropy at each node assumes a given tree. The approach recursively goes down the tree and is given in Algorithm \ref{algo:MCTSentropy_basic}. As mentioned in the paper, the benefit of following the suggested formalization is that it produces the entropy not only of the entire MCTS but also at every node for the subtree spanned from it.

\begin{algorithm}[H]
\caption{MCTS entropy - recursive implementation}\label{algo:MCTSentropy_basic}
\begin{algorithmic}[1]
\Procedure{subTEntropy}{$i, T_i = t$}
\State \% \emph{Initialization:}
\State $localEntropy = 0$
\State $subtreeEntropy = 0$
\State \% \emph{$p[j]$ is the probability that $X_i == \mathcal{A}[j]$:}
\State $p = $ probability vector (index $i$ is $X_i$ given $T_i = t$)
\For{$j = 0$ to $| \mathcal{A}| - 1$}
\State $localEntropy -= p[j] \cdot \log \left( p[j] \right)$
\State $subTreeEntropy += p[j] \cdot$ 
\State subTEntropy( $i+1, T_{i+1} = \{ \mathcal{A}[j], t \}$ )
\EndFor
\State return $localEntropy + subtreeEntropy$
\EndProcedure
\end{algorithmic}
\begin{algorithmic}[1]
\Procedure{MCTSentropy}{MCTStree}
\State $i=0$
\State $t_0 = null$
\State return subTEntropy($i, T_0 = t_0$)
\EndProcedure
\end{algorithmic}
\end{algorithm}


The advantage of the above algorithm is in its simplicity and clarity. Its main disadvantage is that it works for a given finalized probability tree. Thus, if we change something in the original tree (for example, expand by a single node) and wish to receive the updated entropies, using the above algorithm, we will need to re-run it over the entire updated tree. 
The two methods shown for the calculation of the entropy in Algorithm \ref{algo:MCTSentropy_basic} and the Algorithm that piggy-backs MCTS (given in the paper) provide exactly the same values. They are equivalent. 

\section{Proof of Lemma 3} \label{appendix:proofLemma}
\begin{proof}
We first want to show that $\bold{\tilde{p}}$ is a valid probability vector by showing that the sum of its components is one:
\begin{align}
\sum_i \frac{p_i}{1-\hat{p}} - \frac{\hat{p}}{1 - \hat{p}} + \frac{1 - \sum_i p_i}{1 - \hat{p}} & = \frac{ \sum_i p_i - \hat{p} + 1 - \sum_i p_1}{1 - \hat{p}} \nonumber \\
& = \frac{1 - \hat{p}}{1 - \hat{p}} = 1. 
\end{align}
We are comparing two entropies. The original entropy is of $\bold{p}$, and is as follows:
\begin{align} \label{eq:Hp}
H(\bold{p}) = - \sum_i p_i \log p_i - \left(1 - \sum_i p_i\right) \log \left(1 - \sum_i p_i\right) \nonumber
\end{align}
and we want to connect it to the entropy of $\bold{\tilde{p}}$, meaning after the change to the $\ell^{th}$ component. We can do so through the following observations:
\begin{align}
    \bold{p} = & (1- \hat{p}) \cdot \left( \frac{p_1}{1-\hat{p}}, \ldots, \frac{p_{\ell-1}}{1- \hat{p}}, \frac{p_{\ell} - \hat{p}}{1-\hat{p}}, \frac{p_{\ell+1}}{1- \hat{p}}, \ldots, \frac{1- \sum_i p_i}{1- \hat{p}},\right) \nonumber \\
    & + \hat{p} \cdot (0,\ldots,0, 1, 0, \ldots, 0)
\end{align}
Thus, we can think of two random variables. The first is a binary random variable, denoted by $B$, with probability $\hat{p}$ to be one and $1-\hat{p}$ to be zero. The second random variable $X$ is defined over an alphabet of size $|\mathcal{A}|$, and depended on $B$. When $B$ is zero, $X$ is distributed according to 
\begin{align}
    \left( \frac{p_1}{1-\hat{p}}, \ldots, \frac{p_{\ell-1}}{1- \hat{p}}, \frac{p_{\ell} - \hat{p}}{1 - \hat{p}}, \frac{p_{\ell+1}}{1- \hat{p}}, \ldots, \frac{1- \sum_i p_i}{1- \hat{p}},\right).
\end{align}
When $B$ is one $X$ is deterministic and its value is set to the $\ell^{th}$ element in the alphabet. Thus, the entropy can be written as follows:
\begin{align}
    H(\bold{p} ) & = H( X, B) = H( B) + H( X | B) \nonumber \\
    & = H_b(\hat{p}) \nonumber \\
    & +\Pr \left( B = 0 \right) H( X | B = 0) + \Pr\left( B = 1 \right) H( X | B = 1) \nonumber \\ 
    & = H_b(\hat{p}) + (1 - \hat{p}) H( X | B = 0) \nonumber \\
    & = H_b(\hat{p}) + (1 - \hat{p}) H( \bold{\tilde{p}} ).
\end{align}
This concludes the proof. 
\end{proof}

\section{Proof of Theorem 1} \label{appendix:proofTheorem1}
\begin{proof}[Proof of Theorem 1]
When a subtree is removed, the contribution of that subtree reduces to zero. However, the entropy of the tree is now calculated as if this subtree was never part of the tree. The entropy of any subtree is given by (equation (8) in the paper):
\begin{multline} \label{eq:entropyMCTS_recursionGeneral_v2}
H \left( T^i | T_i = t \right) = H \left( X_i | T_i = t \right) + \\ \sum_{j \in \left[ |\mathcal{A}| \right] } \Pr( X_i = x_j | T_i = t) H \left( T^{i+1} | T_i = t, X_i = x_j \right).
\end{multline}
We can identify that the removal has two implications. The first is on the local entropy, meaning the first argument on the right-hand-side. The second is on the proportion of the contributions of the entropies of the child nodes. Both elements change since by removing a subtree we change the probability vector in the parent node from $\bold{p}$ given by equation (20) in the main paper to $\bold{\tilde{p}}$ defined by equation (21) in the main paper. We first write this equation with the following simplified notation:
$\Pr( X_i = x_j | T_i = t) = p_j, \quad \forall j \in |\mathcal{A}|$ 
and
$H \left( T^{i+1} | T_i = t, X_i = x_j \right) = H_j, \quad \forall j \in |\mathcal{A}|.$ 
Thus, the above equation can be written as follows:
\begin{align}
H \left( \tilde{T^i} | T_i = t \right)  
= & H( \bold{\tilde{p}} ) +  \sum_{j \in \left[ |\mathcal{A}| \right], j \neq k } \tilde{p}_j H_j.
\end{align}
where we also simplified the local entropy to a notation that shows the dependency on the probability vector (probability vector is given in bold). 
The change to the local entropy is according to Corollary 4, thus we have 
\begin{align}
H \left( \tilde{T^i} | T_i = t \right)  = & \frac{1}{1- p_k} \left( H(\bold{p}) - H_b(p_k) \right) +  \sum_{j \in \left[ |\mathcal{A}| \right], j \neq k } \tilde{p}_j H_j. \nonumber 
\end{align}
We continue to expand this to obtain the entropy expression of the original tree:
\begin{align}
& H \left( \tilde{T^i} | T_i = t \right)  \\
& = \frac{1}{1- p_k} \left( H(\bold{p}) -H_b(p_k) \right) +  \sum_{j \in \left[ |\mathcal{A}| \right]} \tilde{p}_j H_j - \tilde{p}_k H_k \nonumber \\
& = \frac{1}{1- p_k} \left( H(\bold{p}) -H_b(p_k)\right) +  \sum_{j \in \left[ |\mathcal{A}| \right]} \frac{p_j}{1- p_k} H_j - \frac{p_k}{1- p_k} H_k \nonumber \\
& = \frac{1}{1- p_k} \left( H(\bold{p}) -H_b(p_k) \right) +  \frac{1}{1- p_k} \left( \sum_{j \in \left[ |\mathcal{A}| \right]} p_j H_j - p_k H_k\right) \nonumber \\
& = \frac{1}{1- p_k} \left( H(\bold{p}) -H_b(p_k) +  \sum_{j \in \left[ |\mathcal{A}| \right]} p_j H_j - p_k H_k \right) \nonumber \\
& =  \frac{1}{1- p_k} \left( H(\bold{p}) + \sum_{j \in \left[ |\mathcal{A}| \right]} p_j H_j\right) -\frac{1}{1- p_k} \left(H_b(p_k) + p_k H_k \right) \nonumber \\
& = \frac{1}{1- p_k} \Bigl( H \left( T^i | T_i = t \right) - \left(H_b(p_k) + p_k H_k \right) \Bigr) \nonumber \\
& = H \left( T^i | T_i = t \right) + \frac{p_k}{1- p_k} H \left( T^i | T_i = t \right) \nonumber \\
& - \frac{1}{1- p_k} \left(  H_b(p_k) +  H_k \right) \nonumber \\
& = H \left( T^i | T_i = t \right) + \frac{1}{1- p_k} \left( p_k H \left( T^i | T_i = t \right) -H_b(p_k) -  H_k \right). \nonumber 
\end{align}
Returning to the original notation we reach the desired result.
\end{proof}

\section{Proof of Theorem 2} \label{appendix:proofTheorem2}
\begin{proof}
We consider a parent node with $N_p$ children nodes, and the following probability vector:
\begin{align} \label{eq:parentProbabilityVector}
    \bold{p} = \left( p_1, p_2, \ldots, 1- \sum_i p_i \right)
\end{align}
such that child $i$ has $N_p p_i$ counts. 

We further assume that child $\ell$, which had $N$ counts in its children, has been modified in some way, such that it currently has:
$\tilde{N} = N ( 1 - \tilde{p})$ 
counts, and its entropy has changed from $H_{\ell}$ to $\tilde{H}_{\ell}$. Note that the above change also changes the counts in the parent node:
$\tilde{N}_p = N_p - N \tilde{p}$.
The important connection that we require is that
$N_p p_{\ell} = N + 1$ 
since the fraction that each child contributes, is its own accumulated count at its children plus one (accounting for the node of the child). Given this observation we have that
\begin{align}
    \tilde{N}_p & = N_p - N \tilde{p} = N_p - \tilde{p} ( N_p p_{\ell} - 1) \nonumber \\
    & = N_p \left( 1 - \left(\tilde{p} p_{\ell} - \frac{\tilde{p}}{N_p} \right) \right).
\end{align}
Denoting
\begin{align}
    \hat{p} \equiv \tilde{p} \left( p_{\ell} - \frac{1}{N_p} \right) 
\end{align}
the change in the parent node is according to $\hat{p}$. 
We also have that
\begin{align}
    N(1 - \tilde{p}) + 1 = N_p ( p_{\ell} - \hat{p})
\end{align}
where the left-hand-side is the new counts of the $\ell^{th}$ child. Given this $\bold{\tilde{p}}$ is given according to:
\begin{align}
    \bold{\tilde{p}} = \bigl(\frac{p_1}{1-\hat{p}}, \frac{p_2}{1-\hat{p}}, \ldots, \frac{p_{\ell-1}}{1-\hat{p}}, \frac{p_{\ell} - \hat{p}}{1-\hat{p}}, \frac{p_{\ell+1}}{1-\hat{p}}, \ldots, \frac{1- \sum_{i} p_i}{1-\hat{p}} \bigr).
\end{align}
As a sanity check we can show that summing the elements of $\bold{\tilde{p}}$ equal one. 
Using Lemma 3 this gives us the immediate change to the local entropy. Assuming we know the change of the entropy contribution we can write a first expression of how the entropy at the parent node changes:
\begin{align}
H \left( \tilde{T^i} | T_i = t \right)  
= & H( \bold{\tilde{p}} ) +  \sum_{j \in \left[ |\mathcal{A}| \right], j \neq \ell } \tilde{p}_j H_j + \tilde{p}_{\ell} \tilde{H}_{\ell}.
\end{align}
Thus we have 
\begin{multline}
H \left( \tilde{T^i} | T_i = t \right)  = \frac{1}{1- \hat{p}} \left( H(\bold{p}) - H_b(\hat{p}) \right) +  \\ \sum_{j \in \left[ |\mathcal{A}| \right], j \neq \ell } \tilde{p}_j H_j + \tilde{p}_{\ell} \tilde{H}_{\ell}. 
\end{multline}
We continue to expand this to extract the original expression for the entropy at the parent node:
\begin{align}
& H \left( \tilde{T^i} | T_i = t \right) \\
& = \frac{1}{1- \hat{p}} \left( H(\bold{p}) -H_b(\hat{p}) \right) +  \sum_{j \in \left[ |\mathcal{A}| \right]} \tilde{p}_j H_j + \tilde{p}_{\ell} \left( \tilde{H}_{\ell} - H_{\ell} \right) \nonumber \\
& = \frac{1}{1- \hat{p}} \left( H(\bold{p}) -H_b(\hat{p})\right) +  \sum_{j \in \left[ |\mathcal{A}| \right]} \frac{p_j}{1- \hat{p}} H_j \nonumber \\
& + \frac{p_{\ell} - \hat{p}}{1- \hat{p}} \left( \tilde{H}_{\ell} - H_{\ell} \right) \nonumber \\
& = \frac{1}{1- \hat{p}} \left( H(\bold{p}) -H_b(\hat{p}) \right) \nonumber \\
& +  \frac{1}{1- \hat{p}} \left( \sum_{j \in \left[ |\mathcal{A}| \right]} p_j H_j  + (p_{\ell} - \hat{p}) \left( \tilde{H}_{\ell} - H_{\ell}\right)\right) \nonumber \\
& = \frac{1}{1- \hat{p}} \left( H(\bold{p}) -H_b(\hat{p}) +  \sum_{j \in \left[ |\mathcal{A}| \right]} p_j H_j + (p_{\ell} - \hat{p}) \left( \tilde{H}_{\ell} - H_{\ell} \right) \right) \nonumber \\
& = \frac{1}{1- \hat{p}} \left( H(\bold{p}) + \sum_{j \in \left[ |\mathcal{A}| \right]} p_j H_j\right) \nonumber \\
& -\frac{1}{1- \hat{p}} \left(H_b(\hat{p}) - (p_{\ell} - \hat{p}) \left( \tilde{H}_{\ell} - H_{\ell} \right) \right) \nonumber \\
& = \frac{1}{1- \hat{p}} \Bigl( H \left( T^i | T_i = t \right) - \left(H_b(\hat{p}) - (p_{\ell} - \hat{p}) \left( \tilde{H}_{\ell} - H_{\ell} \right) \right) \Bigr) \nonumber \\
& = H \left( T^i | T_i = t \right) + \frac{\hat{p}}{1- \hat{p}} H \left( T^i | T_i = t \right) \nonumber \\
& - \frac{1}{1- \hat{p}} \left(  H_b(\hat{p}) - (p_{\ell} - \hat{p}) \left( \tilde{H}_{\ell} - H_{\ell} \right) \right) \nonumber \\
& = H \left( T^i | T_i = t \right) \nonumber \\
& + \frac{1}{1- \hat{p}} \left( \hat{p} H \left( T^i | T_i = t \right) -H_b(\hat{p}) +(p_{\ell} - \hat{p}) \left( \tilde{H}_{\ell} - H_{\ell} \right) \right). \nonumber 
\end{align}
Thus, in order to evaluate the change in the entropy of the entire tree we can propagate up the tree, storing at each transition the original entropy and the updated one, so that we know $\tilde{H}_{\ell} - H_{\ell}$, the entropy of the parent node and $p_{\ell}$, meaning the original probability of the changed child, and $\hat{p}$. Note that $\hat{p}$ depends on the counts and thus, we need to know not only the probability vectors and the entropies along the path to the root, but also the counts. 

This concludes the proof.
\end{proof}

\section{Reduction Implementation Additional Aspects} \label{appendix:implementation_reduction}

\begin{figure}[t]
\centering
\begin{subfigure}[t]{0.5\linewidth}
    \centering
    \includegraphics[width=\linewidth]{figures/merge_local_quarterBeta_original_240.pdf}
\end{subfigure}\hfil
\begin{subfigure}[t]{0.5\linewidth}
    \centering
    \includegraphics[width=\linewidth]{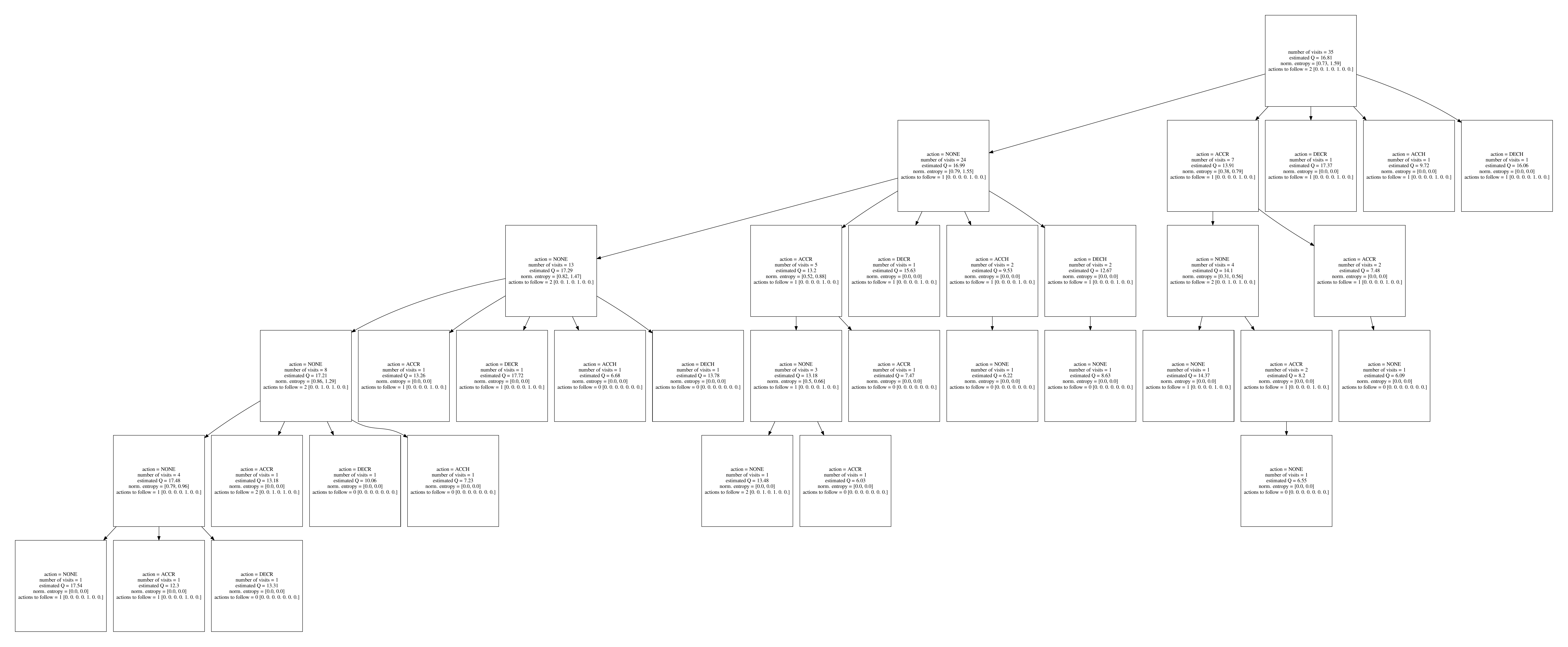}
 \end{subfigure}
 \caption{An example of the reduction using the local algorithm and $\beta = \frac{1}{4} \beta_{UB}$ - merge} \label{fig:example_merge_local}
 \hspace{3cm}
\end{figure}

\begin{figure}[tp]
\centering
\begin{subfigure}[t]{0.5\linewidth}
    \centering
    \includegraphics[width=\linewidth]{figures/curve_two_stage_no_criterion_original_120.pdf}
\end{subfigure}\hfil
\begin{subfigure}[t]{0.5\linewidth}
    \centering
    \includegraphics[width=\linewidth]{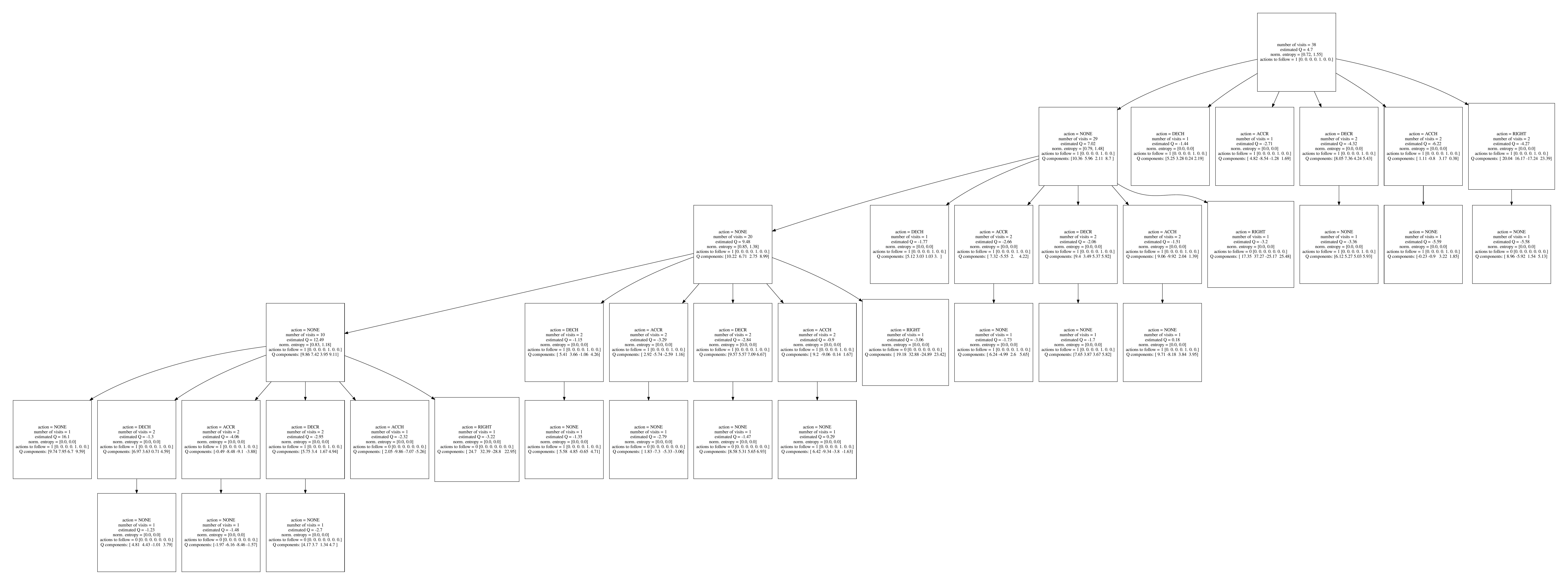}
 \end{subfigure}
 \caption{An example of the reduction using the two-stage algorithm with no criterion (entire $pList$) and $\beta = \frac{1}{4} \beta_{UB}$ - curve \newline} \label{fig:example_curve_two_stage_no_criterion}
 \hspace{3cm}
\end{figure}

As mentioned in the paper we propose two base-line greedy implementations directed by the entropy vs. size trade-off. We show additional examples (on top of Figure 2 and 3 in the paper) of tree reductions in Figures \ref{fig:example_merge_local} and \ref{fig:example_curve_two_stage_no_criterion}. 

We first want to motivate the approach by considering how a local reduction is performed. Fallowing Theorem 1 and Theorem 2 we know that a subtree removal requires the update of the information going up the tree, all the way to the root.  
A basic implementation of such a removal is given in Algorithm \ref{algo:subtreeRemovalBasic}, where $updateEntropy$ is the same method as given in Algorithm 2 in the paper. 

\begin{algorithm}[t]
\caption{Subtree Removal - Basic Implementation}\label{algo:subtreeRemovalBasic}
\begin{algorithmic}[1]
\Procedure{SubtreeRemoval}{$node$}
\State $parent = node.parent$
\State $action = node.action$
\State \% \emph{Initializing a number of visits counter:}
\State $updatedNumberVisits = 0$
\State \% \emph{Removing the pointer to the subtree:}
\State $parent.childrenSummarized.pop(action)$
\State \% \emph{Going up the tree to the root:}
\While{$parent$}
\State \% \emph{Updating the number of visits:}
\State $parent.childNumVisitsSummarized[action] =$ \State $updatedNumberVisits$
\State $updatedNumberVisits =$ \State $sum(parent.childNumVisitsSummarized) + 1$
\State \% \emph{Update the entropy:}
\State updateEntropy($parent$)
\State \% \emph{Going up the tree:}
\State $parent = parent.parent$
\EndWhile
\EndProcedure
\end{algorithmic}
\end{algorithm}

When producing a reduced version we suggest to store the original values as well as the new values, at each node (denoted here as fields with the added $Summarized$).
%
The computational complexity of this update is at most the depth of the tree, meaning logarithmic in the number of nodes.
The issue with the above implementation (Algorithm \ref{algo:subtreeRemovalBasic}) is that it considers a single subtree removal. Thus, using this as a building block suggests that for each subtree that we decide to remove, we update the information up the tree. Given the set of removals from a specific node, this is wasteful in terms of computations. Instead, both suggested algorithm for tree reduction suggest that we perform all chosen removals from a given node in one update up the tree. However, the decision on which subset of subtrees to remove can, by itself, require extensive computations. Specifically, for a criterion such as the trade-off between entropy and size, evaluating each subset also requires going up the tree, meaning logarithmic in the number of nodes of the tree. Thus, in such a case the complexity is $O(N 2^{| \mathcal{A} |} \log N)$. The suggested algorithms can receive different criteria to consider. At each node they locally examine different subsets of nodes to remove, and choose the best subset (according to the given criterion). This forms the basis for both base-line algorithms. The local greedy algorithm immediately continues and performs the removal and updates the tree accordingly.  
The local greedy algorithm gives preference to removals done higher in the tree, regardless of what we have down the tree. The two-stage approach comes to resolve some of that, by placing the best suggested removal from each node in a priority list. The high-level code of the two-stage approach is given in Algorithm \ref{algo:globalSubtreeRemoval}.

\begin{algorithm}[t]
\caption{Two-Stage Subtree Removal}\label{algo:globalSubtreeRemoval}
\begin{algorithmic}[1]
\Procedure{createSummarizedTreeTwoStage}{$tree$, criterion, refValueInit, pListCriterion, updateParentInPlist, $args$}
\State $numberVisits = tree.numberVisits$
\State \% \emph{Initializing values that will propagate down the tree:}
\State $parentsValue = $ -inf
\State $referenceValue =$ refValueInit$(tree, args)$ \label{algo:line:refValueInit}
\State \% \emph{First stage - construction of the priority list:}
\State \% \emph{Initialization:}
\State $pList = priorityList()$ \label{algo:line:init}
\State \% \emph{Calculation of possible reductions down the tree:}
\State reductionCriteria2PrioritizedList($pList, tree,None,$ \label{algo:line:pListCalculation}
\State $\quad$ criterion,updateParentInPlist,
\State $\quad$ $parentsValue$, $referenceValue$, $args$)
\State \% \emph{Post processing of the list:}
\State $pList.$calculateRemianingVisits($numberVisits$) \label{algo:line:postProcessingPlist}
\State \% \emph{Second stage - given the pList reduce the tree:} \label{algo:line:two-stage_highLevelSecondStage}
\State $k = 0$
\While{$k < $len$(pList)$ and pListCriterion($pList, k$)}
\State updateSummarizedTree($pList.retrieveElem(k)$) \label{algo:line:actualUpdating}
\State $k+=1$
\EndWhile
\EndProcedure
\end{algorithmic}
\end{algorithm}

Up to line \ref{algo:line:init} we perform the initialization of the priority list $(pList)$ and different reference values, including a value that is specific to the criterion used (in our case the trade-off of size and entropy) and thus calculated using the input method $refValueInit$ (see line \ref{algo:line:refValueInit}). The actual construction of the prioritized list is done in line \ref{algo:line:pListCalculation} using a recursive method $reductionCriteria2PrioritizedList$ (not detailed). 
Note that this method receives as input two methods that specify how the priority list should be constructed. The first is the actual $criterion$ and the second is how to updated a suggested removal of a parent node assuming one of his descendants is inserted into the priority list - $updateParentInPlist$. Once the priority list is constructed we may perform post-processing of it. 
As an example, we calculated the actual remaining number of visits, by accumulating the proposed removals up to each element in the priority list (line \ref{algo:line:postProcessingPlist}). 
This completes the first stage. The second stage (starting at line \ref{algo:line:two-stage_highLevelSecondStage}) is the actual performance of the suggested removals. This section can also be implemented as shown in Algorithm \ref{algo:line:two-stage_highLevelSecondStage}, which is the version we denoted as V2. Both implementations take into account a condition $pListCriterion$ but the decision whether to stop traversing the list or not is different. 


\begin{algorithm}[H]  
\caption{Two-Stage Subtree Removal - Alternative Second stage (replacing from line \ref{algo:line:two-stage_highLevelSecondStage})} \label{algo:globalSubtreeRemoval_alternativeSecondStage} 
\begin{algorithmic}[1] 
\State \% \emph{Second stage - given the pList reduce the tree:} \label{two-stage_highLevelSecondStage}
\State $k = 0$
\While{$k < $length$(pList)$}
\State $performReduction = $ pListCriterion($pList, k, args$)
\If{$performReduction$}
\State updateSummarizedTree($pList.retrieveElem(k)$) \label{algo:line:actualUpdating}
\EndIf
\State $k+=1$
\EndWhile
\end{algorithmic}
\end{algorithm}

\section{Additional Evaluations} \label{appendix:ExperimentalResults}

We provide here the complete set of results obtained from our experiments comparing the four algorithms (the one-stage algorithm and the three variants of the two-stage algorithm). We begin by providing the exact numerical results calculated over the two driving scenarios examined, the curve highway and ramp-merge scenarios. These are given in Tables 1-4.

\begin{table*}[!h] 
\begin{center} 
\begin{tabular}{ |c |c |c |c |c |c | c | c | c |}
\hline
& Total tree & Main path & Main subtree & $2^{nd}$ path & $2^{nd}$ subtree & Entropy & Trade-off & Number of\\
& reduction	& reduction	& reduction & reduction & reduction & reduction & reduction & trees\\
\hline
$\beta_{UB}$ & $95.82\%$ & $82.04\%$ & $94.89\%$ & $66.30\%$ & $99.12\%$ & $96.27\%$ & $103.66\%$ & $227$ \\
$\frac{1}{2}\beta_{UB}$ & $65.62\%$ & $62.20\%$ & $53.96\%$ & $62.11\%$ & $95.18\%$ & $33.69\%$ & $3.88\%$ & $223$ \\
$\frac{1}{4}\beta_{UB}$ & $61.27\%$ & $47.21\%$ & $61.68\%$ & $4.10\%$ & $63.50\%$ & $6.34\%$ & $-10.24\%$ & $244$ \\
\hline
\end{tabular}
\end{center}
\caption{Performance of the Local Reduction for different values of $\beta$ - merge scenario}
\label{table:Town06_local} 
\end{table*}

\begin{table*}[!h] 
\begin{center} 
\begin{tabular}{ |c |c |c |c |c |c | c | c | c |}
\hline
& Total tree & Main path & Main subtree & $2^{nd}$ path & $2^{nd}$ subtree & Entropy & Trade-off & Number of\\
& reduction	& reduction	& reduction & reduction & reduction & reduction & reduction & trees\\
\hline
$\beta_{UB}$ & $94.26\%$ & $78.96\%$ & $92.24\%$ & $64.98\%$ & $98.31\%$ & $94.23\%$ & $104.79\%$ & $237$ \\
$\frac{1}{2}\beta_{UB}$ & $65.85\%$ & $60.93\%$ & $56.21\%$ & $60.13\%$ & $91.89\%$ & $34.00\%$ & $4.79\%$ & $250$ \\
$\frac{1}{4}\beta_{UB}$ & $61.77\%$ & $48.63\%$ & $61.78\%$ & $4.73\%$ & $66.81\%$ & $6.79\%$ & $-9.94\%$ & $206$ \\
\hline
\end{tabular}
\end{center}
\caption{Performance of the Two-Stage Reduction with \emph{NO} Criterion on the $pList$ for different values of $\beta$ - merge scenario}
\label{table:Town06_two_stage_noCriterion}
\end{table*}

\begin{table*}[!h] 
\begin{center} 
\begin{tabular}{ |c |c |c |c |c |c | c | c | c |}
\hline
& Total tree & Main path & Main subtree & $2^{nd}$ path & $2^{nd}$ subtree & Entropy & Trade-off & Number of \\
& reduction	& reduction	& reduction & reduction & reduction & reduction & reduction & trees \\
\hline
$\beta_{UB}$ & $63.49\%$ & $75.04\%$ & $87.19\%$ & $3.90\%$ & $5.60\%$ & $30.56\%$ & $-89.75\%$ & $250$ \\
$\frac{1}{2}\beta_{UB}$ & $29.88\%$ & $48.33\%$ & $35.73\%$ & $11.76\%$ & $20.45\%$ & $1.04\%$ & $-16.62\%$ & $204$ \\
$\frac{1}{4}\beta_{UB}$ & $9.27\%$ & $20.65\%$ & $9.31\%$ & $0.00\%$ & $15.22\%$ & $-1.48\%$ & $-4.65\%$ & $250$
 \\
\hline
\end{tabular}
\end{center}
\caption{Performance of the Two-Stage Reduction with Criterion on the $pList$ for different values of $\beta$ - merge scenario}
\label{table:Town06_two_stage_criterion} 
\end{table*}

\begin{table*}[!h] 
\begin{center} 
\begin{tabular}{ |c |c |c |c |c |c | c | c | c |}
\hline
& Total tree & Main path & Main subtree & $2^{nd}$ path & $2^{nd}$ subtree & Entropy & Trade-off & Number of\\
& reduction	& reduction	& reduction & reduction & reduction & reduction & reduction & trees\\
\hline
$\beta_{UB}$ & $64.18\%$ & $64.23\%$ & $84.96\%$ & $8.50\%$ & $14.82\%$ & $29.49\%$ & $-75.45\%$ & $250$ \\
$\frac{1}{2}\beta_{UB}$ & $41.39\%$ & $49.08\%$ & $37.96\%$ & $31.23\%$ & $54.91\%$ & $13.99\%$ & $-10.28\%$ & $250$ \\
$\frac{1}{4}\beta_{UB}$ & $34.76\%$ & $23.40\%$ & $32.85\%$ & $0.00\%$ & $33.21\%$ & $-4.18\%$ & $-14.35\%$ & $250$
 \\
\hline
\end{tabular}
\end{center}
\caption{Performance of the Two-Stage Reduction with the alternative traversing over $pList$ with Criterion on the $pList$ for different values of $\beta$ - merge scenario}
\label{table:Town06_two_stageV2_criterion} 
\end{table*}

\begin{figure}
\begin{center}
    \includegraphics[scale=0.5]{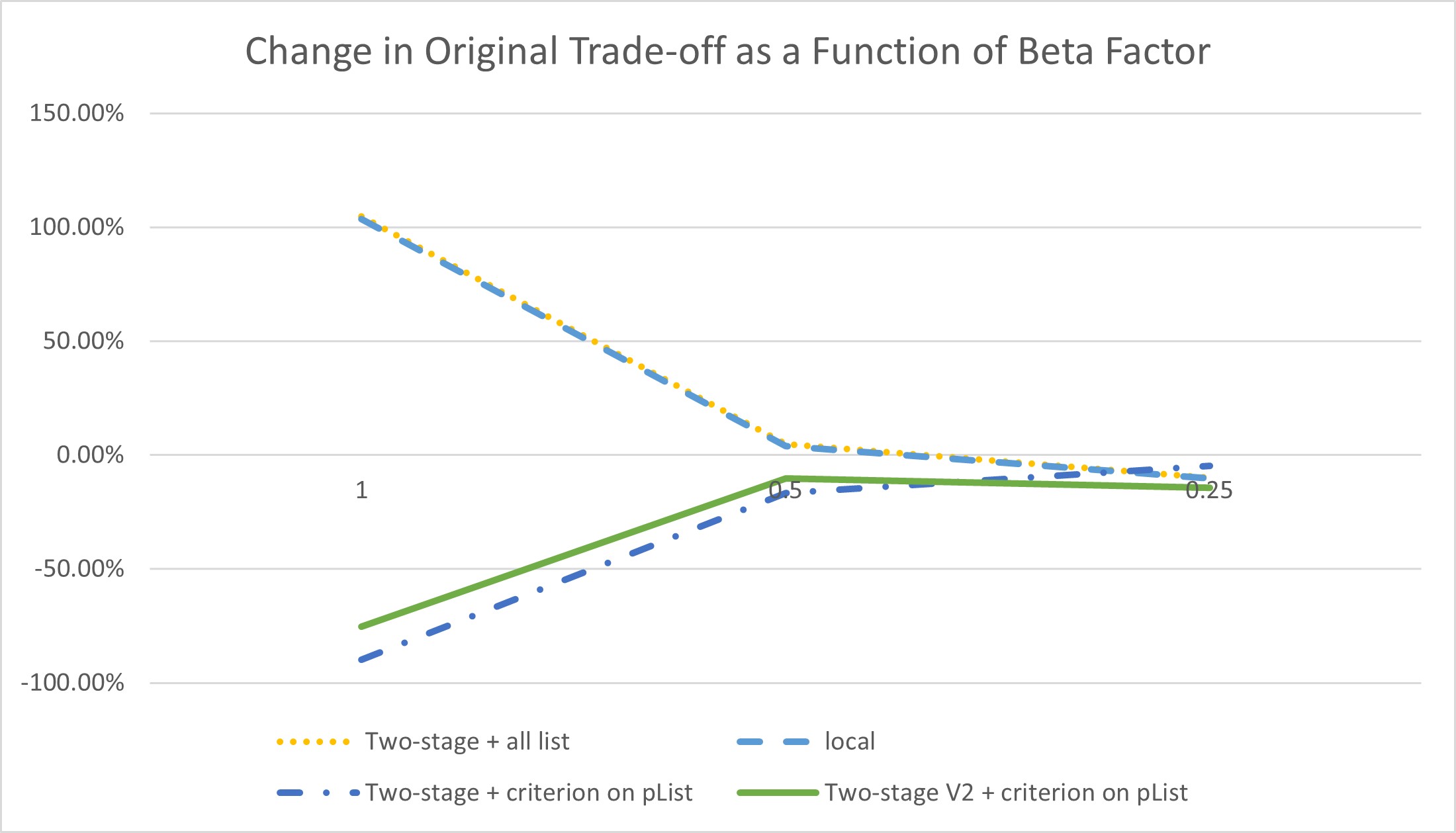}\caption{Comparing the change in trade-off in all four algorithms as a function of the factor of $\beta_{UB}$ - merge scenario \newline} \label{fig:merge_TradeOff} 
\end{center}
\end{figure}

\begin{figure}
\centering
\begin{subfigure}[t]{0.5\linewidth}
    \centering
    \includegraphics[width=\linewidth]{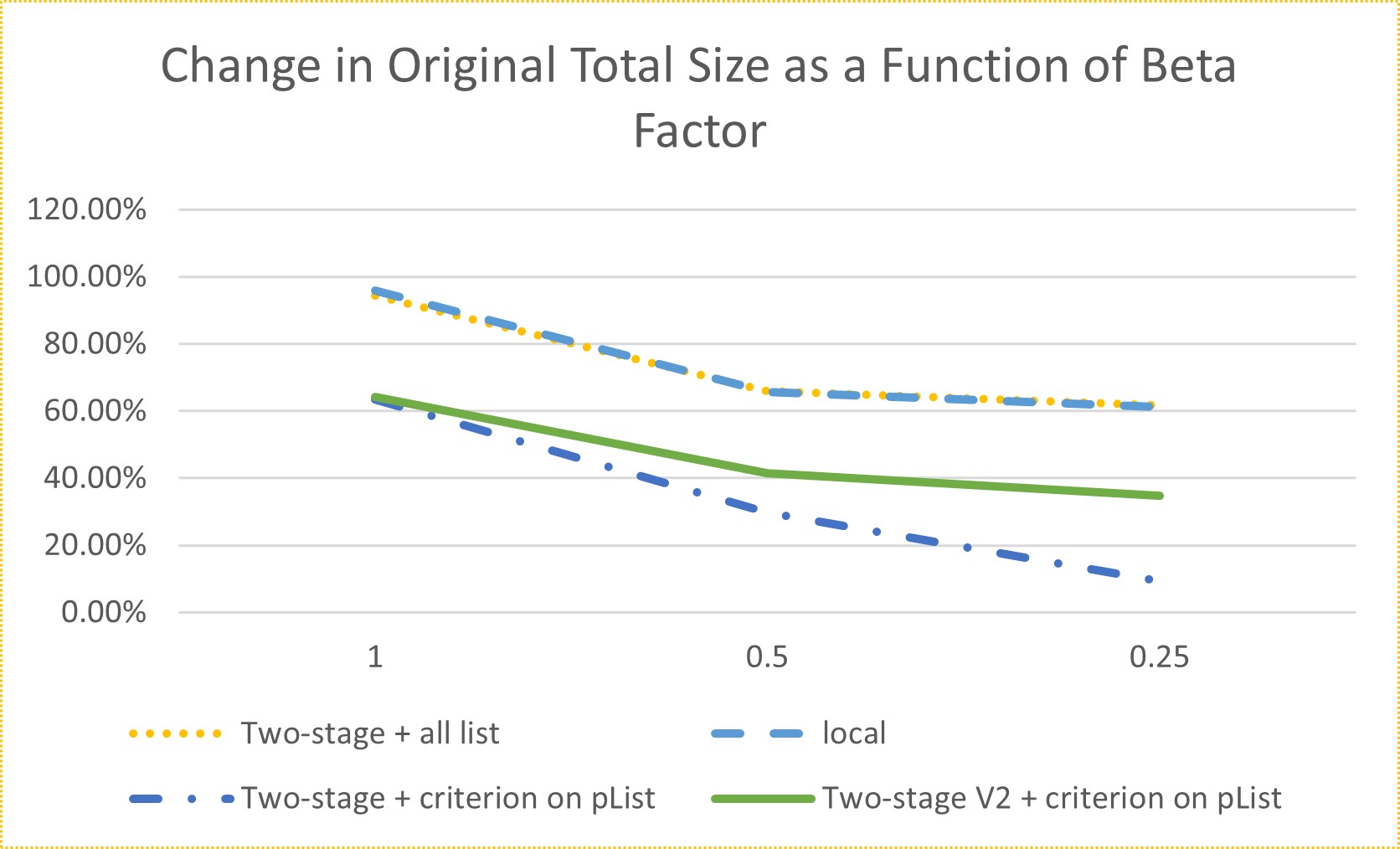}
\end{subfigure}\hfil
\begin{subfigure}[t]{0.5\linewidth}
    \centering
    \includegraphics[width=\linewidth]{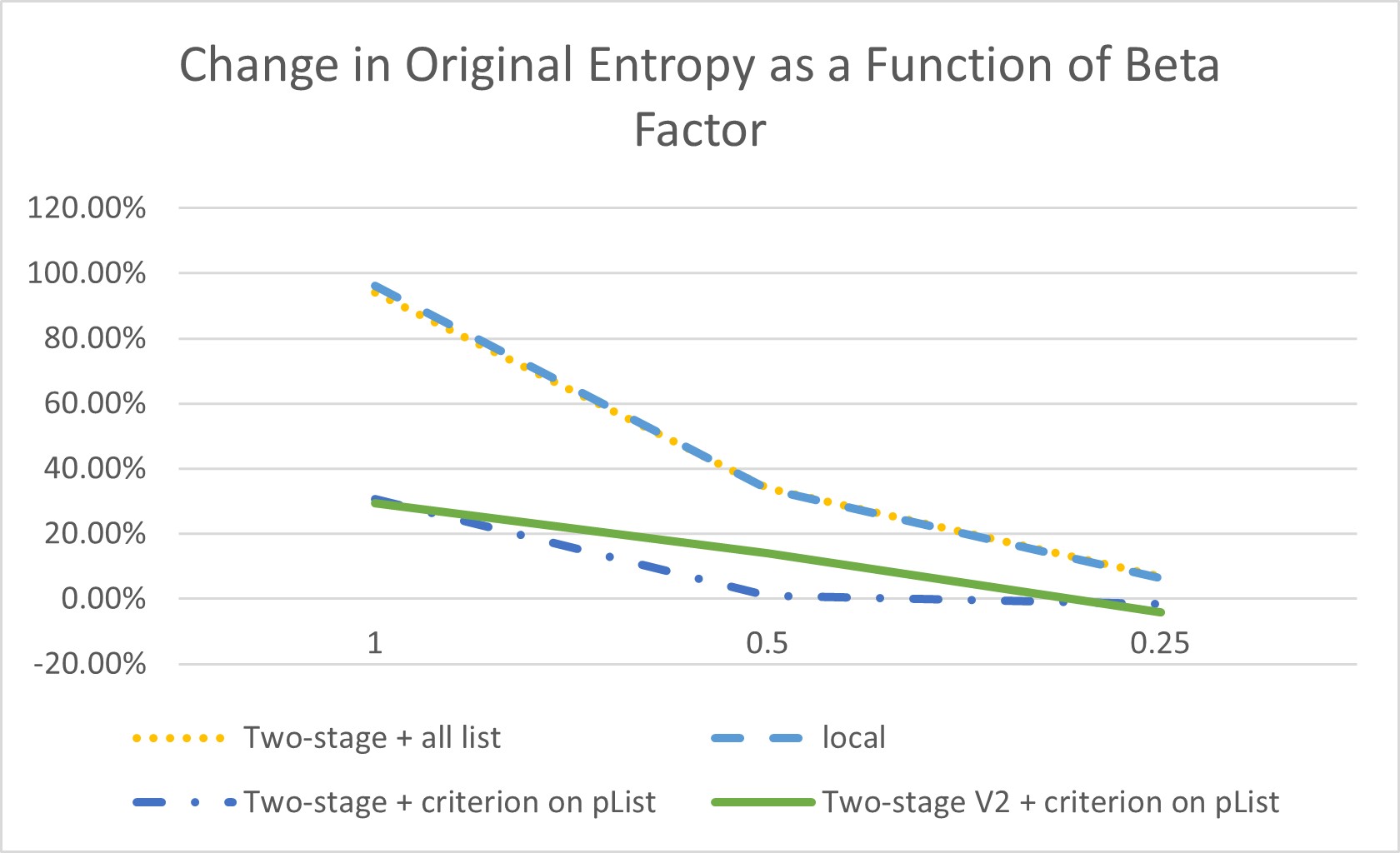}
 \end{subfigure}
 \caption{Comparing the change in total size and entropy 
 in all four algorithms as a function of the factor of $\beta_{UB}$ - merge \newline} \label{fig:merge_SizeAndEntropy}
 \hspace{2cm}
\end{figure}

We also provide here the graphs depicting the main results regarding the influence of the reduction algorithm on the trade-off, entropy and size for the ramp-merge scenario (similar graphs were given in the main text for the highway curve scenario). Figures \ref{fig:merge_TradeOff} and \ref{fig:merge_SizeAndEntropy} summarize the results on the change in trade-off, entropy and size as a function of the factor of $\beta_{UB}$ (equation (13) in the main paper) in the ramp-merge scenario. 


\begin{figure}
\begin{center}
    \includegraphics[scale=0.7]{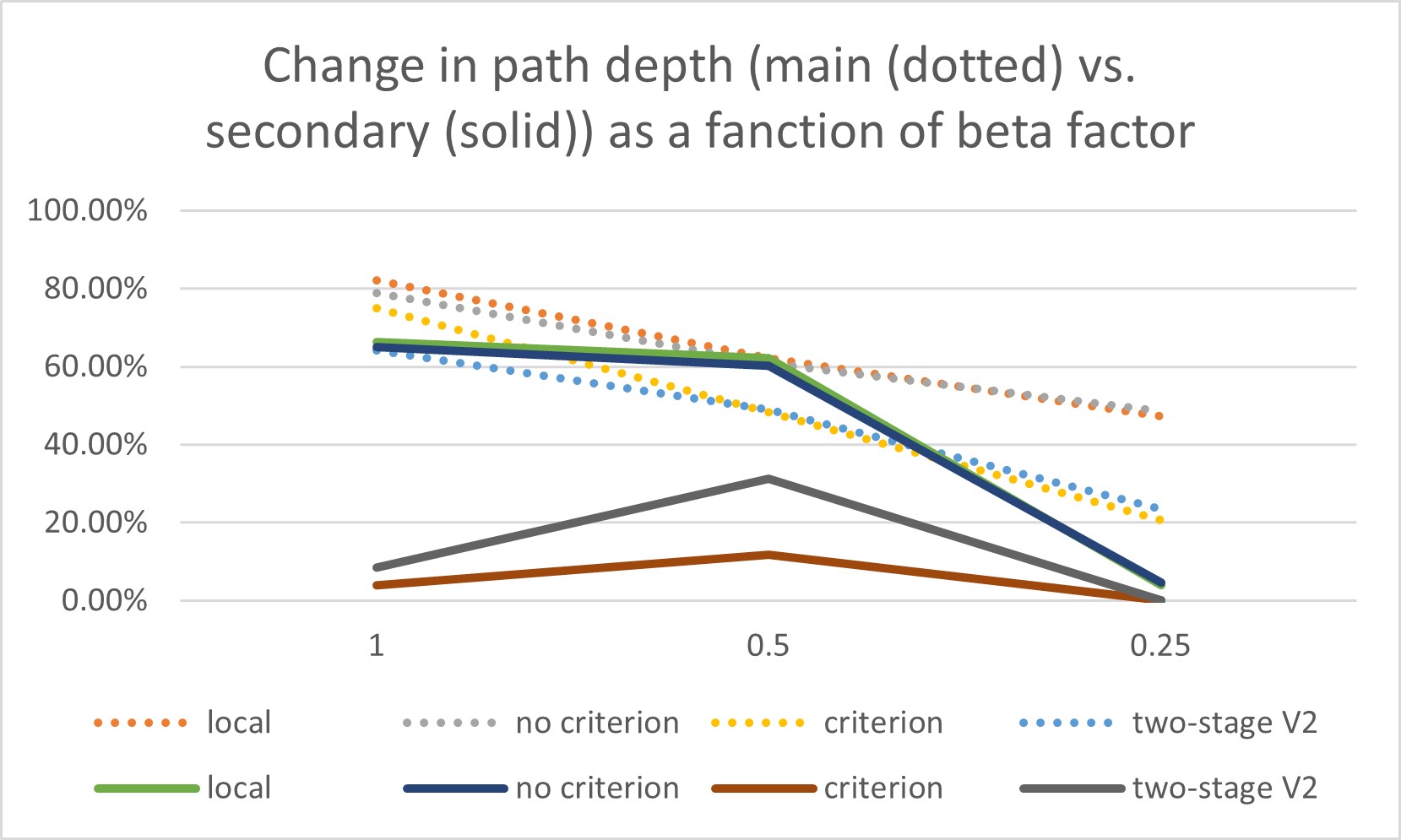}\caption{Comparing the change in depth of the main (dotted) and secondary (solid) paths between the original and reduced tree, as a function of the $\beta$ factor - ramp-merge scenario \newline} \label{fig:merge_depth} 
\end{center}
\end{figure}

\begin{figure}
\begin{center}
    \includegraphics[scale=0.7]{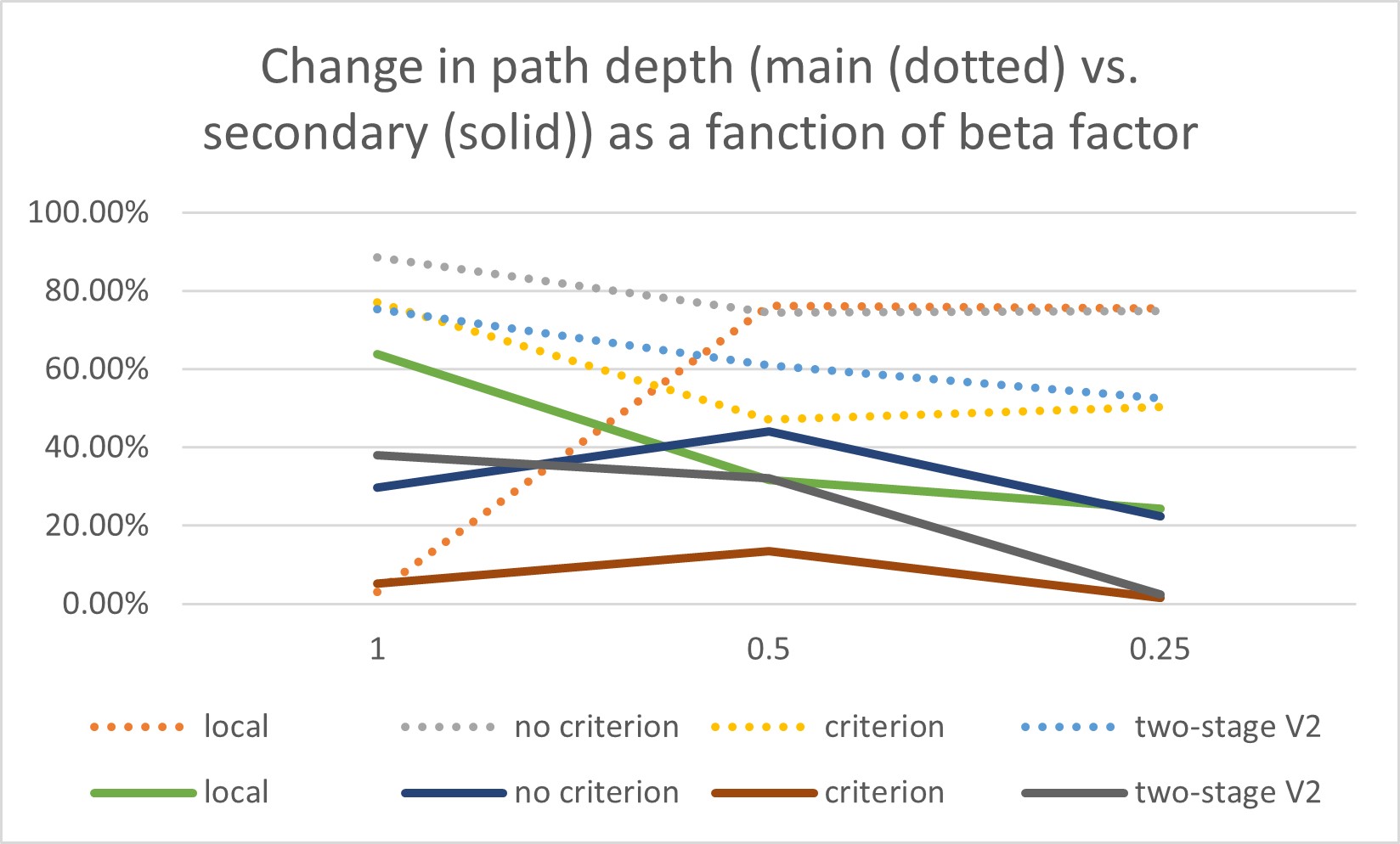}\caption{Comparing the change in depth of the main (dotted) and secondary (solid) paths between the original and reduced tree, as a function of the $\beta$ factor - curve scenario \newline} \label{fig:curve_depth} 
\end{center}
\end{figure}

\begin{figure}
\begin{center}
    \includegraphics[scale=0.7]{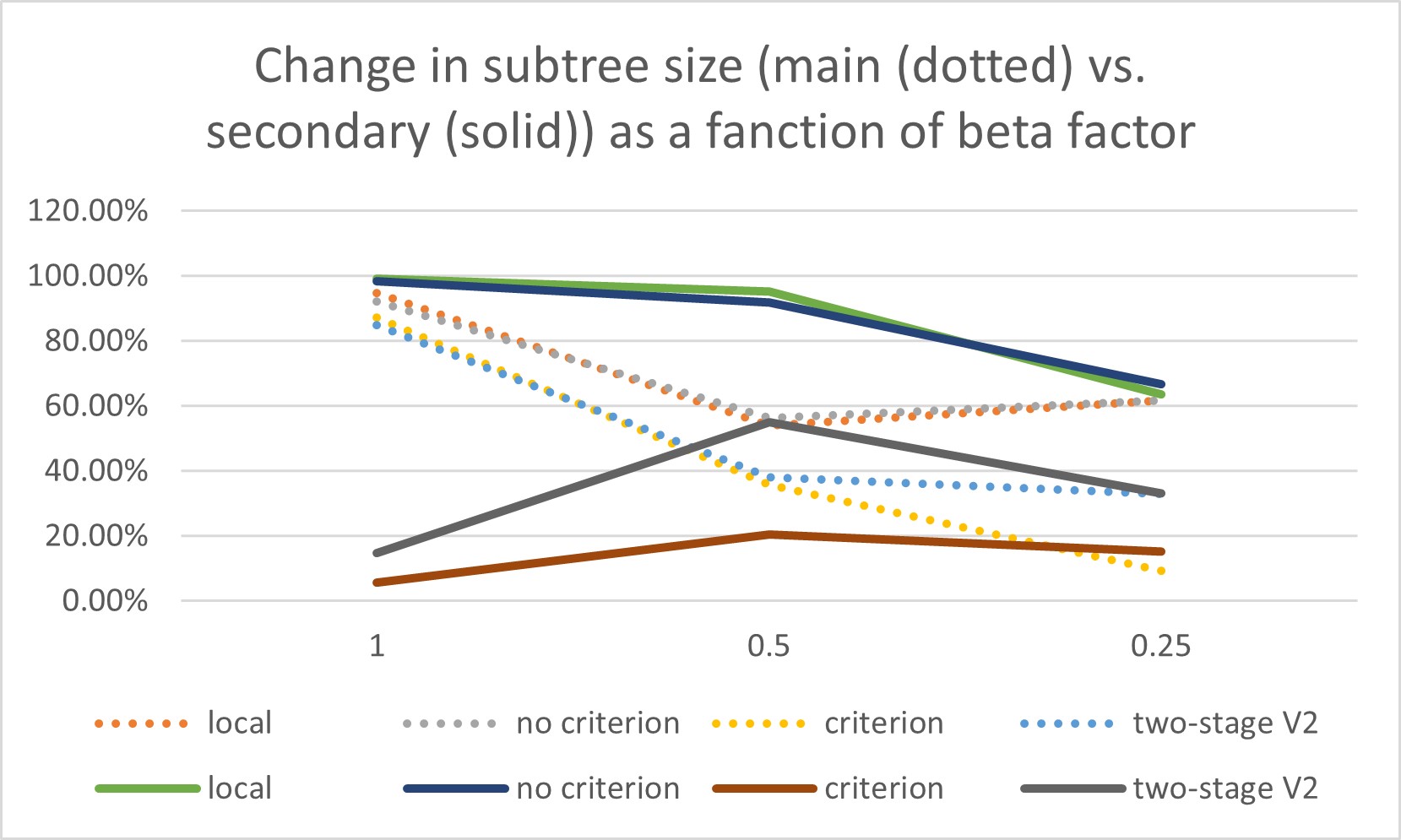}\caption{Comparing the change in size of the main (dotted) and secondary (solid) subtrees between the original and reduced tree, as a function of the $\beta$ factor - ramp-merge scenario \newline} \label{fig:merge_subtree} 
\end{center}
\end{figure}

\begin{figure}
\begin{center}
    \includegraphics[scale=0.7]{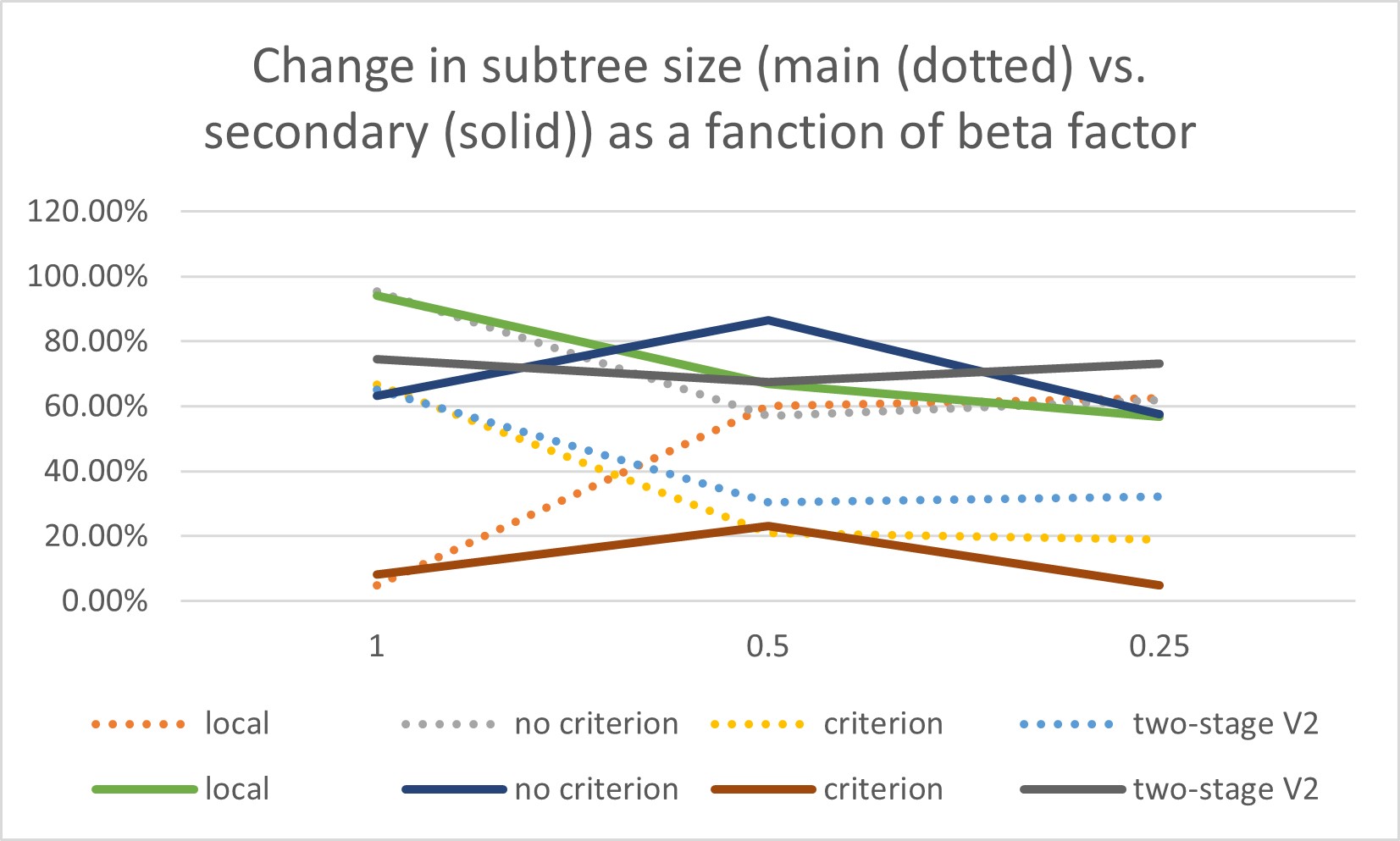}\caption{Comparing the change in size of the main (dotted) and secondary (solid) subtrees between the original and reduced tree, as a function of the $\beta$ factor - curve scenario \newline} \label{fig:curve_subtree} 
\end{center}
\end{figure}

Finally, an important aspect of these reduction algorithms is how they affect the main subtree, and path. To examine this we compared the affect on the main subtree and path vs. the effect on the secondary subtree and path. The effect on the depth of the path for the ramp-merge scenario is shown in Figure \ref{fig:merge_depth} and for the curve scenario in Figure \ref{fig:curve_depth}. The effect on the size of the subtrees for the ramp-merge scanrio is shown in Figure \ref{fig:merge_subtree} and for the curve scenario in Figure \ref{fig:curve_subtree}. We denoted the results on the main path and subtree in dotted lines, whereas those for the secondary path and subtree are denoted in solid lines. The reason for this depiction is that we observe that for the main path and subtree there is usually a convex behavior as a function of $\beta$ whereas in the secondary path and subtree the behavior is usually concave. We found this difference interesting and worth noting. Similar to the overall performance over trade-off, size and entropy, the merge scenario exhibits a clear distinction in performance between the two-stage with criterion algorithms and the local and two-stage with no criterion algorithms, but here the distinction is clear only on the secondary path and subtree. In other words, on the main path and subtree the differences between the four configuration are relatively small, whereas on the secondary path and subtree the difference are considerable. Our explanation for this is that this comes from the built-in mechanism to prefer the preservation of the main path (and subtree) over the secondary one. The algorithm removes the main subtree from a given node only if it chose to remove all children of that node. Thus, we see less difference between the algorithms over the main path and subtree. The secondary path and subtree are influenced more by the specific choice of algorithm, and in the merge scenario this is evident. We clearly see that the two-stage with criterion algorithm reduce less the secondary subtree. Still, we do not see this phenomena in the curve scenario. We believe the explanation for this difference comes from the structure of the trees and is indeed scenario dependent. Examining specific trees like those shown in Figures \ref{fig:example_merge_local} and \ref{fig:example_curve_two_stage_no_criterion} and Figures 2 and 3 in the main text, we see that the curve trees are less balanced and have very small secondary subtrees, while the trees in the merge scenario are slightly more balanced. 

\bibliography{bibXAI}

\begin{thebibliography}{26}
\providecommand{\natexlab}[1]{#1}
\providecommand{\url}[1]{\texttt{#1}}
\expandafter\ifx\csname urlstyle\endcsname\relax
  \providecommand{\doi}[1]{doi: #1}\else
  \providecommand{\doi}{doi: \begingroup \urlstyle{rm}\Url}\fi

\bibitem[Adadi and Berrada(2018)]{AdadiBerrada}
A.~Adadi and M.~Berrada.
\newblock Peeking inside the black-box: A survey on explainable artificial
  intelligence ({XAI}).
\newblock \emph{IEEE Access}, 6:\penalty0 52138--52160, 2018.
\newblock \doi{10.1109/ACCESS.2018.2870052}.

\bibitem[Baier and Kaisers(2020)]{Baier2020TowardsEM}
H.~Baier and M.~Kaisers.
\newblock Towards explainable {MCTS}.
\newblock 2020.
\newblock URL \url{https://api.semanticscholar.org/CorpusID:235602525}.

\bibitem[Beggiato et~al.(2015)Beggiato, Hartwich, Schleinitz, Krems, Othersen,
  and Petermann-Stock]{Beggiato}
M.~Beggiato, F.~Hartwich, K.~Schleinitz, J.~Krems, I.~Othersen, and
  I.~Petermann-Stock.
\newblock What would drivers like to know during automated driving? information
  needs at different levels of automation.
\newblock 11 2015.
\newblock \doi{10.13140/RG.2.1.2462.6007}.

\bibitem[Browne et~al.(2012)Browne, Powley, Whitehouse, Lucas, Cowling,
  Rohlfshagen, Tavener, Perez, Samothrakis, and Colton]{SurveyMCTS}
C.~B. Browne, E.~Powley, D.~Whitehouse, S.~M. Lucas, P.~I. Cowling,
  P.~Rohlfshagen, S.~Tavener, D.~Perez, S.~Samothrakis, and S.~Colton.
\newblock A survey of {Monte Carlo Tree Search} methods.
\newblock \emph{IEEE Transactions on Computational Intelligence and AI in
  Games}, 4\penalty0 (1):\penalty0 1--43, 2012.
\newblock \doi{10.1109/TCIAIG.2012.2186810}.

\bibitem[Chakraborti et~al.(2017)Chakraborti, Sreedharan, Zhang, and
  Kambhampati]{Chakraborti}
T.~Chakraborti, S.~Sreedharan, Y.~Zhang, and S.~Kambhampati.
\newblock Plan explanations as model reconciliation: Moving beyond explanation
  as soliloquy.
\newblock In \emph{Proceedings of the 26th International Joint Conference on
  Artificial Intelligence}, IJCAI'17, page 156–163. AAAI Press, 2017.
\newblock ISBN 9780999241103.

\bibitem[Doshi-Velez and Kim(2017)]{doshivelez2017rigorous}
F.~Doshi-Velez and B.~Kim.
\newblock Towards a rigorous science of interpretable machine learning, 2017.

\bibitem[Goldman and Bustin(2022)]{GoldmanBustin2022}
C.~V. Goldman and R.~Bustin.
\newblock Trusting explainable autonomous driving: Simulated studies.
\newblock In \emph{2022 IEEE Intelligent Vehicles Symposium (IV)}, pages
  1255--1260, 2022.
\newblock \doi{10.1109/IV51971.2022.9827312}.

\bibitem[Go\l{}undefinedbiewski et~al.(2018)Go\l{}undefinedbiewski, Magner, and
  Szpankowski]{EntropyTrees}
Z.~Go\l{}undefinedbiewski, A.~Magner, and W.~Szpankowski.
\newblock Entropy and optimal compression of some general plane trees.
\newblock \emph{ACM Trans. Algorithms}, 15\penalty0 (1), oct 2018.
\newblock ISSN 1549-6325.
\newblock \doi{10.1145/3275444}.
\newblock URL \url{https://doi.org/10.1145/3275444}.

\bibitem[Hoel et~al.(2020)Hoel, Driggs-Campbell, Wolff, Laine, and
  Kochenderfer]{Volvo2019}
C.-J. Hoel, K.~Driggs-Campbell, K.~Wolff, L.~Laine, and M.~J. Kochenderfer.
\newblock Combining planning and deep reinforcement learning in tactical
  decision making for autonomous driving.
\newblock \emph{IEEE TRANSACTIONS ON INTELLIGENT VEHICLES}, 5\penalty0
  (2):\penalty0 294--305, June 2020.

\bibitem[Hoffman et~al.(2018)Hoffman, Mueller, Klein, and
  Litman]{Hoffman2018MetricsFE}
R.~R. Hoffman, S.~T. Mueller, G.~Klein, and J.~Litman.
\newblock Metrics for explainable ai: Challenges and prospects.
\newblock \emph{ArXiv}, abs/1812.04608, 2018.
\newblock URL \url{https://api.semanticscholar.org/CorpusID:54577009}.

\bibitem[Hoffmann and Magazzeni(2022)]{Magazzeni}
J.~Hoffmann and D.~Magazzeni.
\newblock \emph{Explainable AI Planning ({XAIP}): Overview and the Case of
  Contrastive Explanation (Extended Abstract)}, page 277–282.
\newblock Springer-Verlag, Berlin, Heidelberg, 2022.
\newblock ISBN 978-3-030-31422-4.
\newblock URL \url{https://doi.org/10.1007/978-3-030-31423-1_9}.

\bibitem[Kulkarni et~al.(2019)Kulkarni, Zha, Chakraborti, Vadlamudi, Zhang, and
  Kambhampati]{Kulkarni}
A.~Kulkarni, Y.~Zha, T.~Chakraborti, S.~G. Vadlamudi, Y.~Zhang, and
  S.~Kambhampati.
\newblock Explicable planning as minimizing distance from expected behavior.
\newblock In \emph{Proceedings of the 18th International Conference on
  Autonomous Agents and MultiAgent Systems}, AAMAS '19, page 2075–2077,
  Richland, SC, 2019. International Foundation for Autonomous Agents and
  Multiagent Systems.
\newblock ISBN 9781450363099.

\bibitem[Lage et~al.(2019)Lage, Lifschitz, Doshi-Velez, and Amir]{OfraAmir}
I.~Lage, D.~Lifschitz, F.~Doshi-Velez, and O.~Amir.
\newblock Exploring computational user models for agent policy summarization.
\newblock In \emph{Proceedings of the 28th International Joint Conference on
  Artificial Intelligence, IJCAI 2019}, IJCAI International Joint Conference on
  Artificial Intelligence, pages 1401--1407, 2019.
\newblock \doi{10.24963/ijcai.2019/194}.

\bibitem[Lam et~al.(2021)Lam, Lin, Irvine, Dodge, Shureih, Khanna, Kahng, and
  Fern]{lam2021identifying}
K.-H. Lam, Z.~Lin, J.~Irvine, J.~Dodge, Z.~T. Shureih, R.~Khanna, M.~Kahng, and
  A.~Fern.
\newblock Identifying reasoning flaws in planning-based {RL} using tree
  explanations, 2021.
\newblock URL \url{https://arxiv.org/abs/2109.13978}.

\bibitem[Lopez et~al.(2018)Lopez, Behrisch, Bieker-Walz, Erdmann,
  Fl{\"o}tter{\"o}d, Hilbrich, L{\"u}cken, Rummel, Wagner, and
  Wie{\ss}ner]{SUMO2018}
P.~A. Lopez, M.~Behrisch, L.~Bieker-Walz, J.~Erdmann, Y.-P. Fl{\"o}tter{\"o}d,
  R.~Hilbrich, L.~L{\"u}cken, J.~Rummel, P.~Wagner, and E.~Wie{\ss}ner.
\newblock Microscopic traffic simulation using {SUMO}.
\newblock In \emph{The 21st IEEE International Conference on Intelligent
  Transportation Systems}. IEEE, 2018.
\newblock URL \url{https://elib.dlr.de/124092/}.

\bibitem[Nguyen and Mart{\'{\i}}nez(2020)]{nguyen2020quantitative}
A.~Nguyen and M.~R. Mart{\'{\i}}nez.
\newblock On quantitative aspects of model interpretability, 2020.
\newblock URL \url{https://arxiv.org/abs/2007.07584}.

\bibitem[Rosenfeld and Richardson(2019)]{RosenfeldAndRichardson}
A.~Rosenfeld and A.~Richardson.
\newblock Explainability in human–agent systems.
\newblock \emph{Auton Agent Multi-Agent Syst}, 33:\penalty0 673--705, 2019.
\newblock \doi{10.1007/s10458-019-09408-y}.

\bibitem[Sequeira and Gervasio(2020)]{Sequeira_2020}
P.~Sequeira and M.~Gervasio.
\newblock Interestingness elements for explainable reinforcement learning:
  Understanding agents’ capabilities and limitations.
\newblock \emph{Artificial Intelligence}, 288:\penalty0 103367, Nov. 2020.
\newblock ISSN 0004-3702.
\newblock \doi{10.1016/j.artint.2020.103367}.
\newblock URL \url{http://dx.doi.org/10.1016/j.artint.2020.103367}.

\bibitem[Silver et~al.(2017)Silver, Schrittwieser, Karen~Simonyan, Huang, Guez,
  Hubert, Baker, Lai, Bolton, Chen, Lillicrap, Hui, Sifre, van~den Driessche,
  Graepel, and Hassabis]{MasteringGoNature}
D.~Silver, J.~Schrittwieser, I.~A. Karen~Simonyan, A.~Huang, A.~Guez,
  T.~Hubert, L.~Baker, M.~Lai, A.~Bolton, Y.~Chen, T.~Lillicrap, F.~Hui,
  L.~Sifre, G.~van~den Driessche, T.~Graepel, and D.~Hassabis.
\newblock Mastering the game of {Go} without human knowledge.
\newblock \emph{Nature}, 550:\penalty0 354--359, 2017.
\newblock \doi{10.1038/nature24270}.

\bibitem[Silver et~al.(2018)Silver, Hubert, Schrittwieser, Antonoglou, Lai,
  Guez, Lanctot, Sifre, Kumaran, Graepel, Lillicrap, Simonyan, and
  Hassabis]{AlphaGo2018Sceince}
D.~Silver, T.~Hubert, J.~Schrittwieser, I.~Antonoglou, M.~Lai, A.~Guez,
  M.~Lanctot, L.~Sifre, D.~Kumaran, T.~Graepel, T.~Lillicrap, K.~Simonyan, and
  D.~Hassabis.
\newblock A general reinforcement learning algorithm that masters chess, shogi,
  and {Go} through self-play.
\newblock \emph{Science}, 362\penalty0 (6419):\penalty0 1140--1144, 2018.
\newblock \doi{10.1126/science.aar6404}.
\newblock URL \url{https://www.science.org/doi/abs/10.1126/science.aar6404}.

\bibitem[Sreedharan et~al.(2019)Sreedharan, Chakraborti, Muise, and
  Kambhampati]{Sreedharan2019PlanningWE}
S.~Sreedharan, T.~Chakraborti, C.~Muise, and S.~Kambhampati.
\newblock Planning with explanatory actions: A joint approach to plan
  explicability and explanations in human-aware planning.
\newblock \emph{ArXiv}, abs/1903.07269, 2019.
\newblock URL \url{https://api.semanticscholar.org/CorpusID:81981838}.

\bibitem[Sreedharan et~al.(2021)Sreedharan, Chakraborti, and
  Kambhampati]{SREEDHARAN2021103558}
S.~Sreedharan, T.~Chakraborti, and S.~Kambhampati.
\newblock Foundations of explanations as model reconciliation.
\newblock \emph{Artificial Intelligence}, 301:\penalty0 103558, 2021.
\newblock ISSN 0004-3702.
\newblock \doi{https://doi.org/10.1016/j.artint.2021.103558}.
\newblock URL
  \url{https://www.sciencedirect.com/science/article/pii/S0004370221001090}.

\bibitem[Sutton and Barto(2018)]{sutton2018reinforcement}
R.~S. Sutton and A.~G. Barto.
\newblock \emph{Reinforcement learning: An introduction}.
\newblock MIT press, 2018.

\bibitem[\'{S}wiechowski et~al.(2023)\'{S}wiechowski, Godlewski, Sawicki, and
  Ma\'{n}dziuk]{recentModificationsMCTS}
M.~\'{S}wiechowski, K.~Godlewski, B.~Sawicki, and J.~Ma\'{n}dziuk.
\newblock {Monte Carlo Tree Search}: a review of recent modifications and
  applications.
\newblock \emph{Artif Intell Rev}, 56:\penalty0 2497--2562, 2023.
\newblock \doi{10.1007/s10462-022-10228-y}.

\bibitem[Wu et~al.(2017)Wu, Kreidieh, Parvate, Vinitsky, and
  Bayen]{Wu2017FlowAA}
C.~Wu, A.~Kreidieh, K.~Parvate, E.~Vinitsky, and A.~M. Bayen.
\newblock Flow: Architecture and benchmarking for reinforcement learning in
  traffic control.
\newblock \emph{ArXiv}, abs/1710.05465, 2017.
\newblock URL \url{https://api.semanticscholar.org/CorpusID:35869294}.

\bibitem[Zhou et~al.(2021)Zhou, Kamijo, Itoh, and Kitazaki]{ZHOU20211}
H.~Zhou, K.~Kamijo, M.~Itoh, and S.~Kitazaki.
\newblock Effects of explanation-based knowledge regarding system functions and
  driver’s roles on driver takeover during conditionally automated driving: A
  test track study.
\newblock \emph{Transportation Research Part F: Traffic Psychology and
  Behaviour}, 77:\penalty0 1--9, 2021.
\newblock ISSN 1369-8478.
\newblock \doi{https://doi.org/10.1016/j.trf.2020.11.015}.
\newblock URL
  \url{https://www.sciencedirect.com/science/article/pii/S1369847820305830}.

\end{thebibliography}

\end{document}